\pdfoutput=1

\documentclass[11pt]{article}

\usepackage{acl}
\usepackage{tablefootnote}

\usepackage{times}
\usepackage{latexsym}
\usepackage{amsmath}
\usepackage{tabularx}
\usepackage{arydshln}
\usepackage{multirow}
\usepackage{bbm}
\usepackage{booktabs}
\usepackage[pdftex]{graphicx}
\usepackage{makecell}
\usepackage{enumitem}
\usepackage{comment}
\usepackage{caption}
\usepackage{subcaption}
\usepackage[T1]{fontenc}
\usepackage[utf8]{inputenc}
\usepackage{microtype}
\usepackage{inconsolata}

\title{What Matters in Memorizing and Recalling Facts?\\
Multifaceted Benchmarks for Knowledge Probing in Language Models}

\author{
Xin Zhao \\
The University of Tokyo \\
\texttt{xzhao@tkl.iis.u-tokyo.ac.jp} \\\And
Naoki Yoshinaga\qquad\quad Daisuke Oba\thanks{Currently, he works for ELYZA, Inc.} \\
Institute of Industrial Science,\\
The University of Tokyo\\
\texttt{\{ynaga,oba\}@iis.u-tokyo.ac.jp}\\}

\newcommand{\shin}[1]{\textcolor{black}{#1}}
\newcommand{\yn}[1]{\textcolor{black}{#1}}

\begin{document}

\maketitle

\begin{abstract}
\shin{Language models often struggle with handling factual knowledge, exhibiting factual hallucination issue.
This makes it vital to evaluate the models' ability to recall its parametric knowledge about facts.}
In this study, we introduce 
\shin{a knowledge probing benchmark},
BELIEF(-ICL), to evaluate the knowledge recall ability of both encoder- and decoder-based pre-trained language models (PLMs) from diverse perspectives.
BELIEFs utilize a multi-prompt dataset to evaluate PLM's accuracy, consistency, and reliability in factual knowledge recall. 
To enable a more reliable evaluation with BELIEFs, we semi-automatically create MyriadLAMA, which has massively diverse prompts.
\shin{We validate the effectiveness of BELIEFs in comprehensively evaluating PLM's knowledge recall ability on diverse PLMs, including recent large language models (LLMs).
\shin{We then investigate key factors in memorizing and recalling facts in PLMs, such as model size, pretraining strategy and corpora, instruction-tuning process and in-context learning settings.}
Finally, we reveal the limitation of the prompt-based knowledge probing.}
The MyriadLAMA is publicized.\footnote{\url{https://huggingface.co/datasets/iszhaoxin/MyriadLAMA}}
\end{abstract}

\section{Introduction}
One of the strongest motivations for training a language model (LM) using massive text is to increase the ability to handle factual knowledge~\cite{kamalloo-etal-2023-evaluating}.
However, even if LMs are trained on massive text,
they suffer from hallucinations that generate incorrect knowledge-grounded sentences~\cite{zhang2023hallucination}.
Considering that large LMs (LLMs) are being widely applied to real-world tasks, it is vital to evaluate the ability to recall the LLMs' parametric knowledge and what factors influence on memorizing facts during pre-training.


However, evaluating the LLM's knowledge recall ability is still challenging. 
Although LAMA probe~\cite{petroni-etal-2019-language} evaluates the knowledge stored in pre-trained LMs (PLMs), it provides only prediction accuracy.
Some studies diversify prompts in the LAMA probe to compute prediction consistency (robustness)~\cite{elazar-etal-2021-measuring, jiang-etal-2020-know}, but those datasets have either low quality or low quantity issues (\S\ref{sec:dataset_comparision}). 
Moreover, since the LAMA probe assumes encoder-based PLMs with the masked LM objective to solve fill-in-the-blank tasks, directly applying it to decoder-based LLMs will underestimate their knowledge recall ability.
Although recent studies leveraged QA datasets to probe LLMs' knowledge~\cite{kamel2022,mallen-etal-2023-trust,wiland-etal-2024-bear,maekawa-etal-2024-retrieval}, they 
overlook other important aspects than prediction accuracy such as robustness to diverse prompts and the reliability of predictions, which are important for real-world applications.

\begin{figure}
\centering
\setlength{\fboxsep}{90pt}
\includegraphics[width=\linewidth,clip]{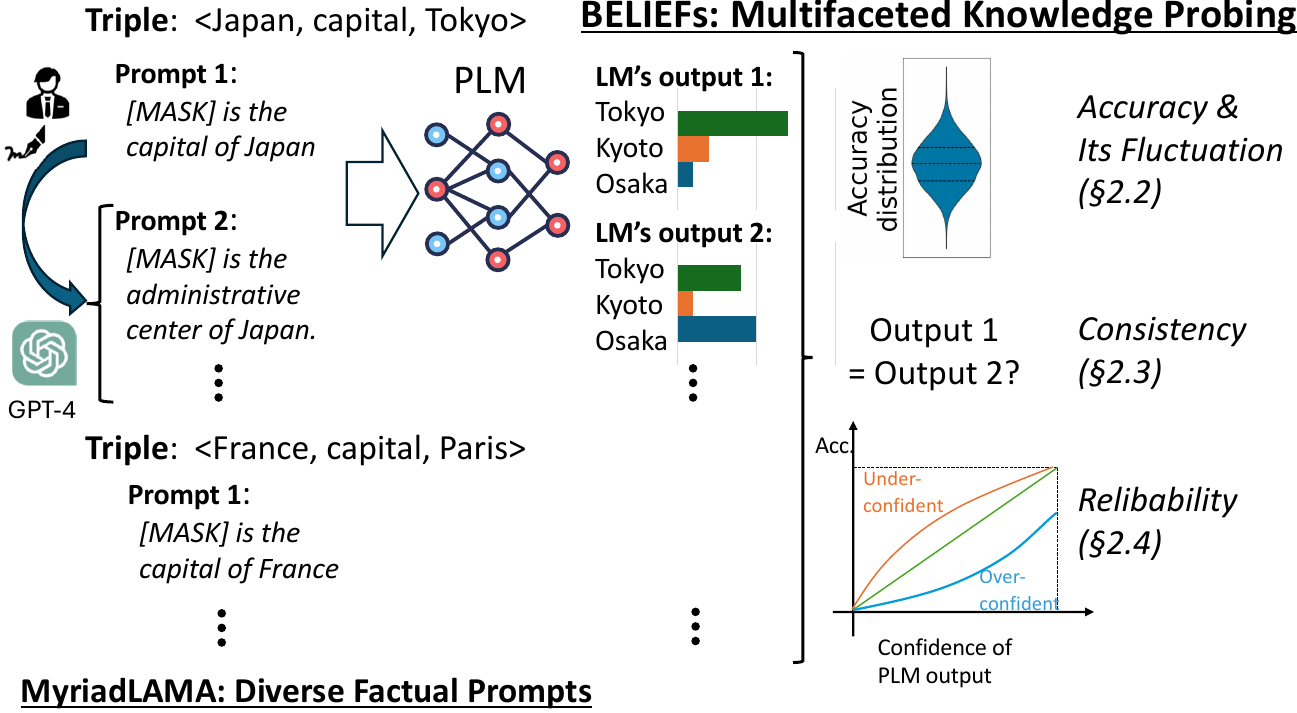}
\caption{BELIEFs with MyriadLAMA: the BELIEF benchmarks utilize diverse factual prompts (here, MyriadLAMA) to assess LM's knowledge recall ability in terms of accuracy, consistency, and reliability.}\label{fig:belief}
\end{figure}

\shin{In this study, we introduce a multifaceted benchmark for knowledge probing, BELIEFs (Figure~\ref{fig:belief}), including BELIEF (\S\ref{sec:belief}) and BELIEF-ICL (\S\ref{sec:belief-icl}) for encoder- and decoder-based PLMs.}
BELIEFs utilize diverse prompts for each fact to account for the impact of linguistic expressions when evaluating LLMs' knowledge recall ability. 
This allows us to evaluate the robustness and reliability of LLM knowledge by measuring fluctuations in accuracy, consistency, and overconfidence in fact prediction.
Since BELIEFs require a multi-prompt probing dataset with diverse prompts for each fact, we build a new probing dataset, MyriadLAMA, to enable a more accurate and comprehensive evaluation (\S\ref{sec:MyriadLAMA}). 
\shin{MyriadLAMA expands LAMA-UHN~\cite{Petroni2020HowCA} by offering different prompts for each fact through a semi-automatic method.}
\shin{Specifically, we obtain a wide variety of lexically, syntactically, and semantically diverse prompts by rewriting relational templates and extending subject expressions.}

We applied BELIEFs to various encoder- and decoder-based PLMs, including BERT~\cite{devlin-etal-2019-bert} and Llama3~\cite{dubey2024llama3herdmodels} (\S\ref{sec:experiments}).
Through extensive evaluations, we verify the utility of BELIEFs in uncovering PLMs' knowledge recall ability (\S\ref{sec:effectiveness}).
Moreover, by comparing different PLMs, we gain insights into the factors affecting knowledge recall of PLMs
\shin{from three aspects: accuracy, reliability and robustness (\S\ref{sec:cross-model}).}

The primary findings in this study are as follows:
\begin{itemize}
\item \shin{
Model size, pretraining strategy, and corpora are crucial factors for memorizing knowledge in LMs during pretraining. 
\item
Whereas instruction-tuning enhances LLMs' ability to follow instructions in BELIEF-ICL, it reduces their knowledge recall ability.}
\item 
The inclusion and selection of demonstrations impact knowledge recall, revealing the gap between memorized and recallable facts.
\item 
Exploring the upper limits of covered knowledge by various methods reveals the limitation of prompt-based knowledge probing (\S\ref{sec:upper-limit}).
\end{itemize}

\section{BELIEF Benchmark} \label{sec:belief}
We first present the multifaceted factual probing benchmark, BELIEF for encoder-based PLMs.
Using a multi-prompt probing dataset, BELIEF evaluates the knowledge recall ability of PLMs from accuracy, robustness and reliability (\S\ref{sec:method:acc}-\ref{sec:method:reliability}).
Here, robustness measures PLMs' ability to maintain consistent accuracy and predictions when given different prompts in evaluation. 
Reliability reflects the extent to which we can trust the PLMs' predictions.


\subsection{Preliminaries}\label{sec:method:pre}
To evaluate the facts in PLMs, BELIEF aggregates results from multiple prompts for each fact to mitigate biases from specific linguistic expressions.
This requires varied expressions for each fact, namely multi-prompt factual probing dataset. 

We assume the fill-in-the-blank settings, where each fact is represented as a \textbf{knowledge triple} $\langle$subject, relation, object$\rangle$ (\textit{e.g.}, $\langle$Tokyo, Capital, Japan$\rangle$).
To probe PLMs for a knowledge triple, we first create a \textbf{masked prompt} (hereafter, \textbf{prompt}) (\textit{e.g.}, ``Tokyo is the capital of [Y]'') for it and then input it into PLMs to see if they correctly predict the object token.
To create such prompts, we first need a template for the relation (hereafter, \textbf{relational template}, \textit{e.g.}, [X] is the capital of [Y]). 
We then fill the template with target knowledge triples, replacing [X] with a subject expression and [Y] with a [\textsc{mask}] token. 
A multi-prompt dataset offers diverse prompts for each fact by providing varied relational templates and entity expressions.

We denote the subject-relation pairs in dataset as $T$, the set of prompts for a given subject-relation pair $t\in{T}$ as $P_t$.
If the output distribution corresponding to mask token of a prompt is $\mathcal{O}=\{(w_j, o_j)|\sum_j o_j=1\}$, the prediction result is defined as the token $\hat{w}=\text{argmax}_{w_j, (w_j, o_j)\in\mathcal{O}} o_j$.

\subsection{Accuracy and its fluctuations}\label{sec:method:acc}
To correctly evaluate the accuracy of PLMs, we aggregate predictions from diverse prompts.
Specifically, we randomly select one prompt for each subject-relation pair $t\in T$ to form a set of prompts for all triples $P=\{p_1, ..., p_{|T|}\}$. 
By feeding these prompts $P$ to PLMs, we can calculate one accuracy value based on their predictions. 
We repeat this process to collect a set of accuracies, which we then use to calculate both average and fluctuation.

\smallskip\noindent\textbf{Average accuracy: }
In BELIEF, accuracy metrics include Acc@1, which measures the rate of prompts with the correct token predicted within the top-$1$ output probabilities. 
Then we repeat this process $N$ times to obtain a set of accuracies, denoted as $V_{\textrm{Acc}@1}$, where $|V_{\textrm{Acc}@1}|=N$.
The final average accuracy is calculated as the mean value of $V_{\textrm{Acc}@1}$. 

\smallskip\noindent\textbf{Fluctuation of accuracy: }
For $V_{\textrm{Acc}@1}$, we evaluate accuracy fluctuations using the range and standard deviation (SD). 
The range is determined by the difference between the maximum and minimum accuracy values in $V_{\textrm{Acc}@1}$.

\subsection{Consistency}
\label{sec:method:consistency}
For each subject-relation pair $t$, we assess the PLM's consistency in predicting the object across different prompts in $P_t$. 
Specifically, we compute the degree of match between the prediction result $\hat{w}_t^i$ for a given prompt $p_t^i$ and the prediction results $\hat{w}_t^j$ for other prompts $p_t^j \in P_t$ (where $j \ne i$), across all subject-relation pairs in $T$:
\begin{align} 
\text{Consist} & = \frac{1}{|T|}\sum_{t\in T} \frac{\sum_{i, j: i \ne j, i, j \leq |{P_t}|}{\mathbbm{1}}{[\hat{w}^t_i=\hat{w}^t_j]}}{\frac{1}{2}|{P_t}|(|{P_t}|-1)}\label{eq:consistency}
\end{align}

\subsection{Reliability}
\label{sec:method:reliability}
The reliability of PLMs reflects the extent to which we can trust the predictions they provide. 
In our study, we measure PLMs' overconfidence level in making fact prediction, drawing from the expected error calibration metric~\cite{desai-durrett-2020-calibration}. 
Specially, we measure the difference between true prediction accuracy and models' confidence to their predicted tokens. 
For each prompt, we first acquire the maximum probability (hereafter, \textbf{confidence}) from the output distribution for the mask token. 
Subsequently, all of the prompts are arranged in descending order based on confidence and segmented into $M$ bins ($P^{(1)}$, $P^{(2)}$, ..., $P^{(M)}$), with the same amount of data points in each bin. 
For each bin $i$, we compute the average accuracy $\overline{\textrm{Acc}@1}^{(i)}$ and average confidence $\overline{o_{max}}^{(i)}$. 
In our work, we use $M=10$ for all the experiments.
Finally, the PLM's overconfidence in predicting the object is assessed by averaging differences between average confidence and accuracy across all bins:
\begin{align} 
\textrm{Ovconf} & = \sum_{i=1}^{M} \frac{|P^{(i)}|}{M} (\overline{o_{max}}^{(i)} - \overline{\textrm{Acc}@1}^{(i)})\label{eq:overconf}
\end{align}

The closer the Ovconf is to zero, the more aligned the model's confidence is with its accuracy, indicating reliable confidence. A negative Ovconf value means the model is underconfident.

\section{BELIEF-ICL for Decoder-based LLMs}
\label{sec:belief-icl}
\yn{Recent} LLMs are
based on decoder-only Transformer architecture, and are trained to predict subsequent tokens in a sequence. This makes it challenging for them to directly predict [MASK] tokens in masked prompts, as they cannot utilize information following the [MASK] (\textit{e.g.}, ``[MASK] and Tokyo are twin cities'').
To comprehensively evaluate LLMs and enable fair comparison between encoder- and decoder-based models, we extend BELIEF to LLMs by employing in-context learning (ICL), termed BELIEF-ICL.

\subsection{In-context learning for fact probe}
The in-context learning ability allows LLMs to perform complex tasks during inference using task-specific prompts~\cite{NEURIPS2020_1457c0d6}.
When designing ICL for evaluating factual knowledge, it is essential to consider \textbf{task instructions} and \textbf{context examples} appended to the target prompts. 

\smallskip\noindent\textbf{1) Task instruction}:
We introduce the mask prediction (MP) instruction for prompting LLMs generating one word answer for the target masked prompt.
The task instruction is formulated as \texttt{``Predict the [MASK] in each sentence in one word.''}.

\smallskip\noindent\textbf{2) Context settings}:
We propose four types of contexts to assess the impact of examplar selection on factual knowledge probing, following the QA format outlined in InstructGPT~\cite{NEURIPS2022_b1efde53}. 
\textbf{zero-shot} uses only instructions; \textbf{X-random} samples X facts from all relations as the few-shot demonstrations; \textbf{X-relation} samples X facts from the same relation but with random templates; \textbf{X-template} samples X facts from the same relations and the same template.

In the few-shot learning settings, we ensure that the target fact is excluded in the examples. 
Refer to \S\ref{sec:prompt_examples} for examples of prompts.

\subsection{Evaluation methods}
Since LLMs generate responses without a token limit, matching the correct answer with the model's output can be challenging. 
Variations in language expressions, such as the presence or absence of articles and singular or plural forms, complicate this process. 
Additionally, the model may generate extra tokens not relevant to the [MASK] token, such as parts of the prompt. 
For example, for the prompt ``John Lennon can play [MASK],'' both ``guitars'' and ``a guitar'' should be considered correct.

To measure BELIEF metrics for LLMs, we compare two strings: the generated text and the correct object expression for Acc@1, and two generated texts for Consist and Ovconf. 
Here, we first normalize strings by tokenizing and lemmatizing them. 
For example, ``a guitar'' and ``guitars'' are normalized to ``a, guitar'' and ``guitar.'' 
If a normalized string list is included in the other (partial matching), they are considered matched.

\smallskip\noindent\textbf{1) Accuracy and its fluctuations}:
Accuracy is calculated by comparing the string generated by the model using a greedy decoding strategy to the correct answers. 
Notably, the matching judgment is one-directional: it only checks if the correct answer is included in the generated string. 
One-directional matching is adopted to avoid incorrect judgments from the model generating unrelated words. 
We use the same $N$ as in \S\ref{sec:method:acc} for accuracy measurement. 

\smallskip\noindent\textbf{2) Consistency}:
We use bi-directional matching to evaluate the consistency (Consist) of generated sequences from two prompts.

\smallskip\noindent\textbf{3) Reliability}:
To calculate overconfidence, we need the model's confidence (probability) in its output. 
However, we cannot obtain this directly from the probability over generated tokens, as LLMs can produce diverse outputs that represent the same answer. 
To address this, we propose an approximate measurement. 
For each prompt, we generate 100 samples using multinomial sampling\footnote{Multinomial sampling selects a next token according to the probability, over the entire vocabulary given by the model.}. We then measure the matching rate between the output generated from greedy decoding and the outputs from the 100 samples. This matching rate serves as the confidence value for the prompt\footnote{This method can approximate the overconfidence calculation in BELIEF of sampling answers from the output distribution. It makes the confidence calculated by BELIEF for encoder-based models comparable to that in BELIEF-ICL.}.
The calculation of Ovconf follows the same setting in \S\ref{sec:method:reliability}. 
This method can approximate the BELIEF's Ovconf calculation as BELIEF sampling answers from the output distribution.
Note that, due to the high cost of generating 100 samples for each fact, we adopt a more efficient approach. We sample 10K prompts from 10K unique subject-relation pairs and only use these 10K prompts for answer sampling.

\section{MyriadLAMA Dataset}\label{sec:MyriadLAMA}

The fairness and accuracy of BELIEF evaluation depend on the diversity and quality of multi-prompt factual probing datasets. However, existing datasets are either manually rewritten in small numbers~\cite{elazar-etal-2021-measuring} or mined from texts~\cite{jiang-etal-2020-know}.
The former is accurate but lacks diversity, providing an average of 7.3 prompts per fact with limited variation. 
For example, templates like ``[X] works as [Y]'' and ``[X], who works as [Y]'' are provided as different templates but very similar.
Additionally, the number of templates is highly imbalanced, with 8 out of 46 relations having only one template, while P138\footnote{\url{https://www.wikidata.org/wiki/Property:P9138}} has 20. 
The latter is diverse but includes templates that do not necessarily imply the relationship. 
For instance, for relation P937 (work location)\footnote{\url{https://www.wikidata.org/wiki/Property:P937}}, the mined templates include ``[X] to meet [Y].,'' which significantly deviates from the original meaning.
To achieve a more accurate and fair evaluation, we introduce MyriadLAMA, a new multi-prompt factual probing dataset with improved diversity \yn{while retaining} quality.
Refer to \S\ref{sec:dataset_comparision} for detailed \yn{qualitative and quantitative comparisons} between MyriadLAMA and prior datasets.


\subsection{Dataset construction}
We build MyriadLAMA by semi-automatically extending the existing single-prompt probing dataset LAMA-UHN~\cite{Petroni2020HowCA}.
MyriadLAMA generates multiple prompts for each fact by providing multiple, equal relational templates for each relation and varying the linguistic expressions of subjects. Additionally, MyriadLAMA offers multiple expressions for each object to cover missed facts that are correctly predicted but in different tokens. For example, for the query ``John Lennon was born in [MASK]'', acceptable tokens could include ``UK'' and ``Britain.''\footnote{We follow the setting of LAMA-UHN triples where the object is a single token according to the BERT tokenizer. During evaluation, we consider the fact to be present, if the model's predicted token matches any of the correct tokens, regardless of which correct answer is predicted.}


Specifically, we define knowledge triples that neglect the diversity of surface expressions as \textbf{unique triples} and distinguish them from \textbf{derived triples}, which embody the diverse entity expressions and relational templates in each unique triple.
The triple extension methods are described below. 

\smallskip\noindent\textbf{Extending entities:}
The knowledge triples in LAMA-UHN constitute a subset of the Wikipedia knowledge base T-REx ~\cite{elsahar-etal-2018-rex}. 
T-REx selectively includes only certain objects for subject-relation pairs.
MyriadLAMA extends the unique triples in LAMA-UHN by mining T-REx using subject-relation as key to include other available objects.
For example, if LAMA-UHN contains only E\_\{guitar\} for instruments that ``John Lennon'' can play, we can extend the unique triple to include E\_\{piano\}.
We also extend the entity expressions using aliases obtained from Wikidata.\footnote{\url{https://www.wikidata.org/wiki/Wikidata:Data_access}}

\smallskip\noindent\textbf{Paraphrasing relational templates:}
MyriadLAMA creates a great variety of relational templates by a semi-automatic process. 
Firstly, we manually generate five distinct templates for each relation. 
They incorporate entailment expressions and diverse syntactic patterns like statements and question-answer formats to provide semantic and syntactic variations.
Next, to enhance quantity and lexical diversity, we automatically paraphrase each manually created template 19 times using the GPT-4 API\@.\footnote{OpenAI: gpt-4-1106-preview} Finally, all templates are filtered by human reviewers to remove low quality templates, yielding a total of 4100 templates covering 41 relations.

\subsection{Dataset Statistics}

\begin{table}[t]
\centering
\small
\tabcolsep 5.5pt
\begin{tabular}{lrr}
\toprule
& \textbf{LAMA-UHN} & \textbf{MyriadLAMA} \\ \midrule
Relational templates    & 41 & 4100 \\
Unique triples    & 27,106 & 34,048 \\
Derived triples   & 27,106 & 21,140,500 \\
Subject-relation pairs   & 24,643 & 24,643 \\
Prompts      & 24,643 & 6,492,800 \\\bottomrule
\end{tabular}
\caption{Statistics of LAMA-UHN and MyriadLAMA.}
\label{tab:sample1}
\end{table}

Table~\ref{tab:sample1} lists the statistics of MyriadLAMA\@.
The number of derived triples is increased from 27,106 in LAMA-UHN to 21,140,500, by combining various semi-automatically generated relational templates and the alias expressions for subject and object entities.
As the prompts are generated from derived triples without considering the object expressions, the number of generated prompts are less than the number of derived triples, which is increased from 24,643 to 6,492,800. 
Refer to the appendices for details on dataset construction (\S\ref{sec:dataset_creation}) and validity analysis of MyriadLAMA (\S\ref{sec:dataset_evaluation}). 
Examples of extended templates are provided in \S\ref{sec:template_examples}.


\section{Effectiveness of BELIEFs}
\label{sec:effectiveness}


\subsection{Experimental setups}
\label{sec:experiments}

We use BELIEFs to evaluate the knowledge recall abilities of both encoder- and decoder-based PLMs. 
The target encoder-based PLMs include BERT$_\mathrm{base}$, BERT$_\mathrm{large}$, and BERT$_\mathrm{wwm}$.\footnote{BERT$_\mathrm{wwm}$ masks all tokens for a single word at the same time, while BERT$_\mathrm{base}$ and BERT$_\mathrm{large}$ masks a single token.}
\shin{The target decoder-based LLMs include Llama2 (7B, 13B, and 70B) and Llama3 (8B and 70B), without and with instruction tuning (except for Llama3-70B),
along with Phi3 (mini, small, and medium).
Their brief pretraining information are listed in Table~\ref{tab:llm_info2}.}
Refer to \S\ref{sec:experiment-details} for more details.


\begin{table}[t]
\small
\centering
\tabcolsep 0.5pt
    \begin{tabular}{lrl}
    \toprule
    \multirow{2}{*}[-2pt]{\textbf{PLMs (\#params)}} & 
    \multicolumn{2}{c}{\textbf{Pre-training corpora}} \\ 
    \cmidrule(lr){2-3}
    & \multicolumn{1}{c}{\textbf{size}} & \textbf{source} \\
    \midrule
    BERT$_\mathrm{base}$    (110M) & 
    \multirow{3}{*}[1.2pt]{$\left.\begin{array}{@{}l@{}}\text{3.3B+}\\ \text{3.3B+} \\ \text{3.3B+}\end{array}\right\rbrace$} & 
    \multirow{3}{*}{\makecell[l]{English Wikipedia\\ \& BookCorpus}}\\
    BERT$_\mathrm{large}$  (336M) & \\
    BERT$_\mathrm{wwm}$    (336M) & \\ \midrule
     Llama2-7B\shin{(-IT)}       (7B)  &    \multirow{3}{*}[1.2pt]{$\left.\begin{array}{@{}l@{}}\text{2.0T}\\ \text{2.0T} \\ \text{2.0T}\end{array}\right\rbrace$} &  
     \multirow{3}{*}{\makecell[l]{A collection of publicly\\ available online data.}} \\
     Llama2-13B\shin{(-IT)}    (13B) & & \\
     Llama2-70B\shin{(-IT)}    (70B) &  & \\ \midrule
     Llama3-8B\shin{(-IT)}      (8B)  &     \multirow{2}{*}[1.2pt]{$\left.\begin{array}{@{}l@{}}\text{15T+}\\ \text{15T+}\end{array}\right\rbrace$} 
 & \multirow{2}{*}{\makecell[l]{A collection of publicly\\ available online data.}} \\
     Llama3-70B     (70B) & \\ \midrule
     Phi3-mini       (3.8B) & \multirow{2}{*}[1.2pt]{$\left.\begin{array}{@{}l@{}}\text{4.9T}\\ \text{4.9T}\\ \text{4.9T}\end{array}\right\rbrace$} & 
     \multirow{3}{*}{\makecell[l]{High-quality educational\\ 
     data/code/chat \& synthetic\\ textbook-like data}} \\ 
     Phi3-small     (7B) &  \\ 
     Phi3-medium   (14B) &  \\ 
     \bottomrule
    \end{tabular}
    \caption{The PLMs for evaluation.}
    \label{tab:llm_info2}
\end{table}

We conduct a full-scale evaluation on LLMs with up to 8 billion parameters. 
To save cost of LLM inference, we use five manually rewritten templates only for the LLMs with more than 8B parameters, including Llama2-70B and its IT variant Llama2-70B-IT, Llama3-70B, and Phi3-medium.\footnote{The partial evaluation is sufficient to compare performance across different model sizes.}
To calculate the average and fluctuation of accuracy (\S\ref{sec:method:acc}), we set a large sample number ($N=50,000$) to provide stable, accurate result.

In the following sections, we analyze the evaluation results on various PLMs to deepen our understanding of how PLMs learn and represent factual knowledge. 
All evaluation results, including those for another family of encoder-based models, ALBERT, are presented in Section \S\ref{sec:full-result}.

\begin{table}[t]
    \small
    \centering
    \tabcolsep 1.4pt
    \begin{tabular}{llcccccc}
        \toprule
        \multicolumn{2}{c}{\multirow{3}{*}[2pt]{\textbf{PLMs}}} & 
        \multicolumn{2}{c}{\textbf{Acc@1}} & 
        \multicolumn{2}{c}{\textbf{Fluctuation$\downarrow$}} & 
        \multirow{2}{*}[-2pt]{\textbf{Consist $\uparrow$}} & 
        \multirow{2}{*}[-2pt]{\textbf{Ovconf}} \\ 
        \cmidrule(lr){3-4}
        \cmidrule(lr){5-6}
        & & \textbf{LU} & \textbf{MyL} & \textbf{range} & \textbf{SD} & &\\
        \midrule
        \multirow{3}{*}[4pt]{\rotatebox{-90}{BERT}}
        & BERT$_\mathrm{base}$ & .2403 & .1095 & .1534 & .0217 & .1682 & .2154 \\ 
        & BERT$_\mathrm{large}$ & \textbf{.2454} & .1102 & .1574 & .0220 & \textbf{.1713} & .2052 \\
        & BERT$_\mathrm{wwm}$ & .2448 & \textbf{.1364} & \textbf{.1517} & \textbf{.0208} & .1524 & \textbf{.1000} \\ \midrule
        \multirow{4}{*}[7pt]{\rotatebox{-90}{Llama3-8B}}
         & zero-shot      & .3708 & .3427 & .2864 & .0350 & .0240 & -.1119 \\
         & 4-random       & .5050 & .5205 & .2033 & .0273 & .2156 & -.0789 \\
         & 4-relation     & n/a\tablefootnote{X-relation cannot be applied to single-prompt dataset.} & .6871 & .1236 & .0156 & .3659 & -.0783 \\
         & 4-template     & \textbf{.6490} & \textbf{.7268} & \textbf{.0220} & \textbf{.0026} & \textbf{.4015} & \textbf{-.0582} \\
        \bottomrule
    \end{tabular}
    \caption{\yn{Evaluation results of BERT variants (above) and Llama3-8B with BELIEF/-ICL (below).} \textbf{LU} indicates \textbf{L}AMA-\textbf{U}HN evaluation results, and \textbf{MyL} represents the average Acc@1 on \textbf{My}riad\textbf{L}AMA.}
    \label{tab:single_prompt_eval_result}
\end{table}

\subsection{Do BELIEFs provide additional insights?}
\begin{figure}[t]
    \centering
    \includegraphics[width=\linewidth]{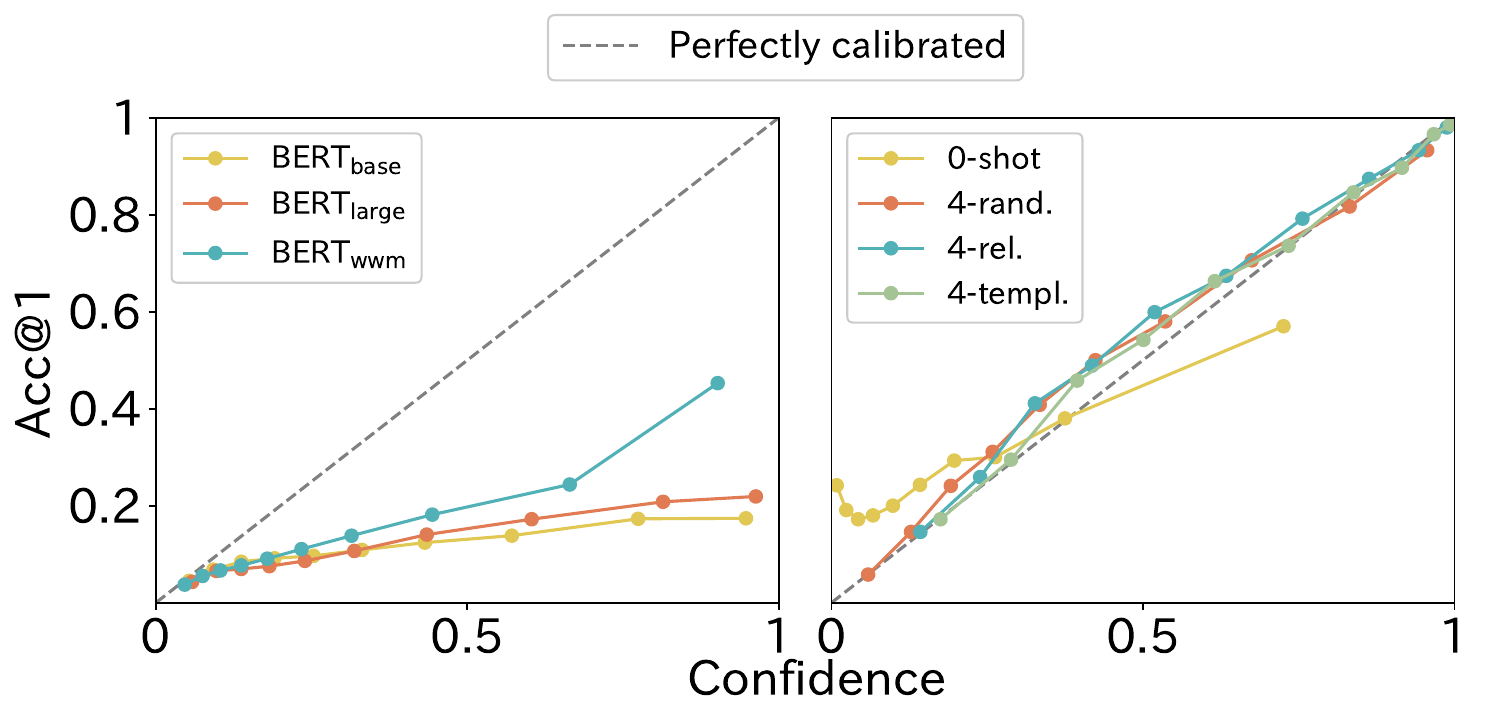}
    \caption{Calibration between confidence and Acc@1. (\textbf{Left}: Three BERT models types; \textbf{Right}: Llama3-8B with four different ICL settings.)}
    \label{fig:Overconf}
\end{figure}

BELIEFs offer evaluation from diverse perspectives rather than accuracy. 
As shown in Table~\ref{tab:single_prompt_eval_result} (Above), the evaluation result highlights accuracy fluctuations among the BERT variants. 
All BERT models show low consistency and tend to be overconfident in their predictions. 
Figure~\ref{fig:Overconf} (left) depicts the relationship between confidence and Acc@1 of the BERT models, indicating low accuracy even for prompts with \yn{confident outputs}. 
Whereas  BERT$_\mathrm{wwm}$ performs better over most BELIEF metrics, BERT$_\mathrm{large}$ outperforms BERT$_\mathrm{wwm}$ on LAMA-UHN\@. 
This discrepancy arises from the limited prompts used in LAMA-UHN and the 
\yn{the single-faceted}
evaluation method.
This highlights BELIEF's effectiveness in achieving a more accurate factual probing comparison between PLMs.

\subsection{Does ICL adhere to instructions?}
We then explore the effectiveness of different ICL settings in extracting facts from LLMs.  
We evaluate the instruction adherence of these settings from two aspects: predicting facts and generating one-word answers, reflecting that the target objects in MyriadLAMA are primarily one-word entities.

\shin{Table~\ref{table:instruct_follow} shows Acc@1 and one-word generation ratio of two pretrained LLMs (Llama2-7B and Llama3-8B) and one instruction-tuned LLM (Phi3-small).} 
We found that under few-shot settings, even the pretrained LLMs exhibit a remarkable ability to follow instructions, indicating the effectiveness of prompting LLMs to predict mask tokens through in-context learning.
Our evaluation with QA-style ICL settings also confirms this (see \S\ref{sec:qa-icl} for details). 
Moreover, exemplars similar to the target prompt (4-template) in the context boosted improvement overall metrics (Table~\ref{tab:single_prompt_eval_result} \textbf{Below}, Table~\ref{table:instruct_follow}).

\begin{table}[t]
\small
\centering
    \begin{tabular}{lcc}
    \toprule
    \multirow{3}{*}[7pt]{\textbf{ICL settings}} & 
    \textbf{\makecell{Fact prediction\\(Acc@1)}} & 
    \textbf{1-word ratio} \\ 
    \cmidrule(lr){2-3}
    & \multicolumn{2}{c}{\textbf{\makecell{(Llama2-7B / Llama3-8B / Phi3-small)}}}\\
    \midrule
     zero-shot      & .3385/.3427/.4258 & .4802/.1572/.8883 \\
     4-random       & .4816/.5205/.4889 & .8058/.8147/.8913 \\
     4-relation     & .6286/.6871/.6339 & .9246/.9071/.9287 \\
     4-template     & \textbf{.6616}/\textbf{.7268}/\textbf{.6612} & \textbf{.9266}/\textbf{.9187}/\textbf{.9411} \\
     \bottomrule
    \end{tabular}
    \caption{Instruction following rates on Llama2-7B, Llama3-8B and Phi3-\yn{small}.}
    \label{table:instruct_follow}
\end{table}

\subsection{Can BELIEFs mitigate bias?}
We explore whether BELIEFs can mitigate prompt bias in evaluations. 
To measure prompt bias quantitatively, we use content-free prompts, where the subject is replaced by meaningless tokens~\cite{Zhao2021CalibrateBU, xu-etal-2024-take-care}, and collect the probabilities of candidate tokens in the output distributions over the mask token.\footnote{Specifically, we adopt a similar setting to ~\cite{Zhao2021CalibrateBU}, by ensembling the distribution over prompts with three content-free tokens: ``N/A,'' an empty string, and ``?''.}
We measure the bias level of the prompt using the certainty of distributions over candidate tokens. 
Specifically, we define bias level as follows:
\begin{align}
\text{bias-level} &= 1 - \frac{\mathcal{H}}{\mathcal{H}_{max}}
\end{align}
where $\mathcal{H}$ is the entropy, and $\mathcal{H}_{max}$ is the maximum entropy for the uniform distribution with same size.

We measure bias in both single- and multi-prompt evaluations. In single-prompt evaluation, we represent bias as the average level across all relational templates. 
For measuring bias-level in multi-prompt evaluation, we first average output distributions by different templates for each relation, then use the bias-level of the averaged distribution to quantify it.
Taking P31:instance-of\footnote{\url{https://www.wikidata.org/wiki/Property:P31}} as an example, the average probability of ``science'' over all templates is 8.30\%, but it rises to 52.79\% for template: ``[Y] contains [X] as one of its elements.''

\section{Differentiating PLMs in Fact Probing}
\label{sec:cross-model}
This section compares the PLMs' knowledge recall abilities in terms of accuracy, reliability, and robustness and then explores factors affecting them.

\subsection{Factors affecting the recall accuracy}
\smallskip\noindent\textbf{1) Pre-training strategy.}
Table~\ref{tab:single_prompt_eval_result} confirms that BERT$_\mathrm{wwm}$ outperforms BERT$_\mathrm{large}$ in terms of all metrics, while BERT$_\mathrm{wwm}$ differs from BERT$_\mathrm{large}$ only in the masking strategy during pre-training.
The superiority of BERT$_\mathrm{wwm}$ likely stems from its challenging pre-training paradigm, which requires recalling whole words without sub-token information, enhancing word-level contextual understanding. 
This underscores the importance of pre-training strategy in knowledge acquisition.

\begin{table}[t]
\small
\centering
\tabcolsep 2.4pt
    \begin{tabular}{lccccc}
    \toprule
    \multirow{2}{*}[-2pt]{\textbf{PLMs}} & 
    \multirow{2}{*}[-2pt]{\textbf{Acc@1}} & 
    \multicolumn{2}{c}{\textbf{Fluctuation}$\downarrow$} &
    \multirow{2}{*}[-2pt]{\textbf{Consist}$\uparrow$} & 
    \multirow{2}{*}[-2pt]{\textbf{Ovconf}} \\
        \cmidrule(lr){3-4} & 
        \textbf{} &\textbf{range} & \textbf{SD} \\
    \midrule
     Llama2-7B      & .6699 & .0257 & .0034 & .4174 & -.0933 \\
     Llama2-13B     & .7080 & .0235 & .0031 & .4326 & \textbf{-.0662} \\
     Llama2-70B     & \textbf{.7784} & \textbf{.0190} & \textbf{.0024} & \textbf{.4449} & -.0690 \\
     \midrule
     Llama2-7B-IT   & .6013 & .0368 & .0045 & .3629 & .2007 \\
     Llama2-13B-IT  & .6482 & .0301 & .0038 & .3656 & .1708 \\
     Llama2-70B-IT  & \textbf{.7232} & \textbf{.0258} & \textbf{.0031} & \textbf{.4226} & \textbf{.1026} \\
     \midrule
     Llama3-8B    & .7316 & .0194 & .0025 & .4060 & -.1119 \\
     Llama3-70B   & \textbf{.8211} & \textbf{.0139} & \textbf{.0017} & \textbf{.4636} & \textbf{-.0812} \\
     \midrule
     Phi3-mini     & .6106 & .0314 & .0039 & .3686 & .0911 \\
     Phi3-small    & .6668 & .0306 & .0039 & .3667 & .1221 \\
     Phi3-medium   & \textbf{.7100} & \textbf{.0207} & \textbf{.0025} & \textbf{.4009} & \textbf{.0317} \\
     \bottomrule
    \end{tabular}
    \caption{BELIEF-ICL evaluation on LLMs with the 4-template setting using manually-rewritten templates.}
    \label{table:belief_icl_diff_sizes}
\end{table}

\smallskip\noindent\textbf{2) Model size.} 
\label{paragraph:model_size}
Table~\ref{table:belief_icl_diff_sizes} compares the knowledge recall abilities of LLMs with difference sizes.\footnote{Owing to the high computational cost of inference on large LLMs like Llama2-70B, we select only five manually rewritten templates with 4-template ICL setting for evaluation.}
We can observe that larger LLMs consistently achieve higher accuracy in predicting facts.
Combining with the improvement from BERT$_\mathrm{base}$ to BERT$_\mathrm{large}$ from Table~\ref{tab:single_prompt_eval_result}, 
the importance of model size in fact acquisition during pre-training is confirmed.

\begin{table}[t]
\small
\centering
\tabcolsep 2.8pt
    \begin{tabular}{lccccc}
    \toprule
    \multirow{2}{*}[-2pt]{\textbf{PLMs}} & 
    \multirow{2}{*}[-2pt]{\textbf{Acc@1}} & 
    \multicolumn{2}{c}{\textbf{Fluctuation$\downarrow$}} &
    \multirow{2}{*}[-2pt]{\textbf{Consist$\uparrow$}} & 
    \multirow{2}{*}[-2pt]{\textbf{Ovconf}}\\
    \cmidrule(lr){3-4} & 
    \textbf{} &\textbf{range} & \textbf{SD} & &\\
    \midrule
     \shin{Llama2-7B-IT}   & .2925 & \textbf{.1980} & \textbf{.0253} & .1151 & .2605 \\
     \shin{Llama3-8B-IT}   & .3578 & .2213 & .0262 & .1660 & \textbf{.1402} \\
     Phi3-mini      & \textbf{.4258} & .2437 & .0292 & \textbf{.1782} & .2171 \\
     \bottomrule
    \end{tabular}
    \caption{BELIEF-ICL evaluation with zero-shot setting on instruction-tuned LLMs.}
    \label{table:zero-shot-llms}
\end{table}

\smallskip\noindent\textbf{3) Pre-training corpora.} 
Table~\ref{table:belief_icl_diff_sizes} shows that Llama3-8B outperforms larger Llama2-13B in fact probing. This is likely due to Llama3's pre-training corpus being seven times larger than Llama2 (Table~\ref{tab:llm_info2}). 
Meanwhile, Llama3-70B surpasses Llama2-70B, confirming the importance of pre-training data volume for fact acquisition.

In the zero-shot evaluation using the entire MyriadLAMA, as shown in Table~\ref{table:zero-shot-llms}, 
\shin{Phi3-mini outperforms Llama2-7B-IT and Llama3-8B in knowledge retrieval. 
Given that Phi3-mini (3.8B) has about half the model size of Llama2-7B-IT and Llama3-8B-IT, and model size typically enhances knowledge retrieval, this result is notable.}
This superior performance can be attributed to the high-quality, textbook-like material used for pre-training the Phi3 models, highlighting the significant impact of high-quality training data.

\smallskip\noindent\textbf{\shin{4) Instruction-tuning.}} 
\shin{Table~\ref{tab:pretrained_vs_instruction_tuned} confirms that instruction-tuned Llama2-7B-IT exhibit a higher one-word generation rate than Llama2-7B, as expected. 
However, the instruction-tuned LLM consistently demonstrate lower Acc@1 scores on different ICL settings. 
This indicates a potential negative impact of instruction-tuning on the models, where general language understanding can improve, but some factual knowledge is partially lost as a result of the tuning process.}

\smallskip\noindent\textbf{\shin{5) Inclusion and selection of demonstrations.}} As shown in Table~\ref{tab:pretrained_vs_instruction_tuned} and Table~\ref{table:experiment-partial}, using demonstrations in prompts consistently improves Acc@1. 
Including few-shot demonstrations with same templates to the target question can nearly double Acc@1 values (from zero-shot to 4-template settings). 
Closer demonstrations also enhance performance across all metrics, highlighting a significant gap between the factual knowledge LLMs memorized and what they can actually recall.

\begin{table}[t]
    \small
    \centering
    \tabcolsep 1.2pt
    \begin{tabular}{llcccccc}
        \toprule
        \multicolumn{2}{c}{\multirow{3}{*}[2pt]{\textbf{PLMs}}} & 
        \multirow{2}{*}[-2pt]{\textbf{Acc@1}} & 
        \multicolumn{2}{c}{\textbf{Fluctuation$\downarrow$}} & 
        \multirow{2}{*}[-2pt]{\textbf{Consist$\uparrow$}} & 
        \multirow{2}{*}[-2pt]{\textbf{Ovconf}} & 
        \multirow{2}{*}[-2pt]{\textbf{\makecell{1-word\\ratio}}} \\ 
        \cmidrule(lr){4-5}
        & & & \textbf{range} & \textbf{SD} & &\\
        \midrule
        \multirow{4}{*}[5pt]{\rotatebox{-90}{\scriptsize{Llama2-7B}}}
         & 0-shot      & .3385 & .2602 & .0299 & .1269 & -.1119 & .4752\\
         & 4-rand.       & .4816 & .2250 & .0270 & .2312 & \textbf{-.0894} & .8247\\
         & 4-rel.     & .6286 & .1221 & .0150 & .3753 & -.1335   & .9060\\
         & 4-templ.     & \textbf{.6616} & \textbf{.0294} & \textbf{.0036} & \textbf{.4163} & -.0933 & \textbf{.9299}\\
         \midrule
         \multirow{4}{*}[7pt]{\rotatebox{-90}{\scriptsize Llama2-7B-IT}}
         & 0-shot      & .2925 & .1980 & .0253 & .1151 & .2605  & .9069\\
         & 4-rand.       & .4334 & .1958 & .0229 & .2128 & .2410 & .9081 \\
         & 4-rel.     & .5576 & .0791& .0092 & .3341 & \textbf{.1900} & .9314\\
         & 4-templ.     & \textbf{.5896} & \textbf{.0439} & \textbf{.0050} & \textbf{.3687} & .2061 & \textbf{.9380}\\
        \bottomrule
    \end{tabular}
    \caption{\shin{Evaluation results of pretrained Llama2-7B (above) and instructed-tuned Llama2-7B-IT (below).}}
    \label{tab:pretrained_vs_instruction_tuned}
\end{table}

\subsection{Factors affecting the reliability}
\label{sec:analysis_reliability}

Table~\ref{tab:single_prompt_eval_result} shows a significant difference in Ovconf between BERT models and Llama3-8B, with BERT models being overconfident and Llama3-8B being underconfident.
In this section, we explore the reasons for these differences and investigate additional factors affecting reliability beyond model size.

\smallskip\noindent\textbf{1) The number of output tokens.} 
One main difference in Ovconf calculation between encoder- and decoder-based PLMs is that the decoder-based PLMs will generate multiple tokens.
Thus, we investigate the effect of output token count on Ovconf values. We divide the MyriadLAMA prompt set into groups based on the number of tokens generated.
For each group, we calculate the probability of the entire token sequence and compute Ovconf for token counts from 1 to 5.\footnote{The prompts generated within five tokens cover 98.78\% of Llama3-8B's generations with the 4-template ICL setting.}

The Ovconf values for each group \yn{of 1 to 5 output tokens} on Llama3-8B (4-template) are -0.1030, -0.0906, -0.0297, -0.0546, and 0.0573, showing models become more overconfident with more output tokens.
This trend is consistent across models.


\begin{figure}[t]
    \centering
    \includegraphics[width=\linewidth]{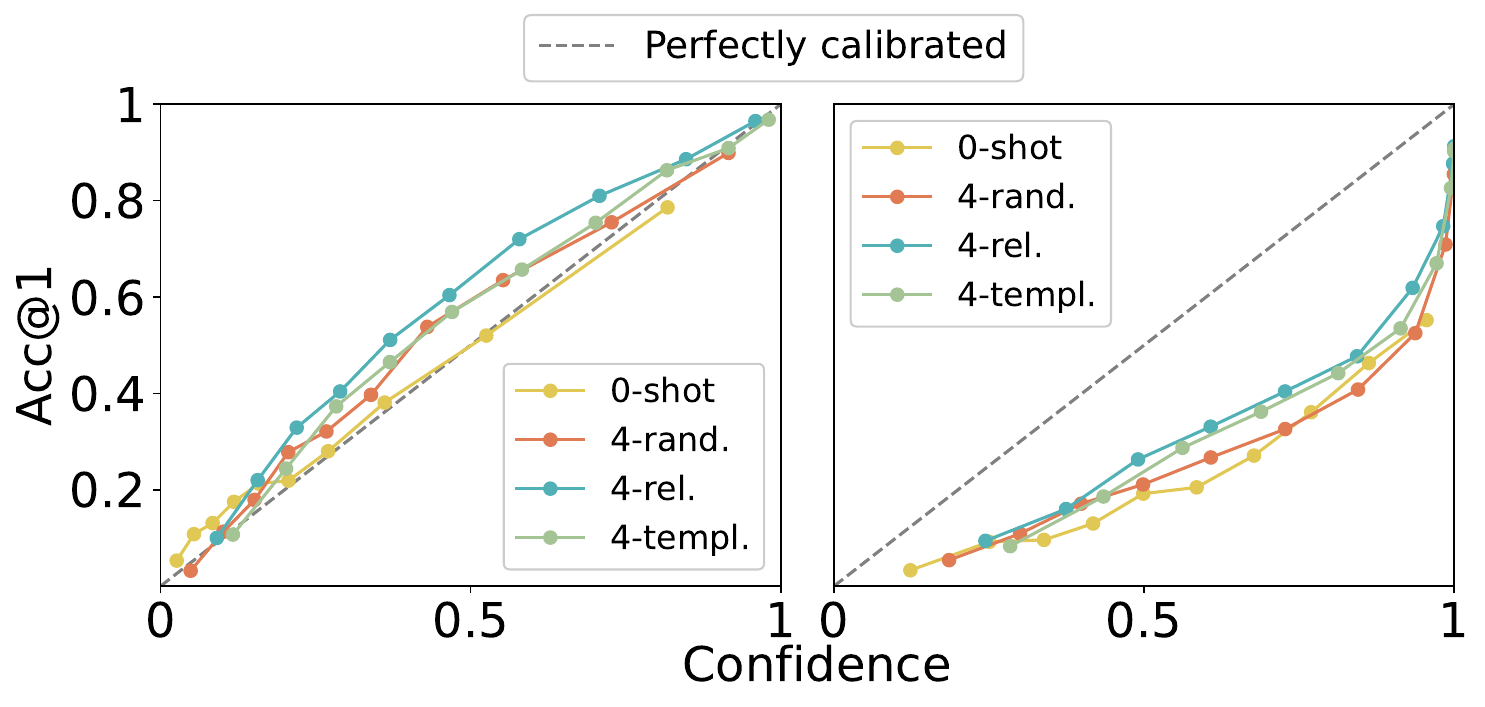}
    \caption{Calibration plots of Llama2-7B (left) and Llama2-7B-IT (right).)}
    \label{fig:Ovconf_llama2_vs_it}
\end{figure}

\smallskip\noindent\textbf{\shin{2) \shin{Instruction-tuning inflates LLMs confidence}}.}
\shin{Table~\ref{tab:pretrained_vs_instruction_tuned} and Figure~\ref{fig:Ovconf_llama2_vs_it} confirm that instruction-tuned LLMs make the models overly confident in their outputs.
The pretraining use more diverse language data with uncertainties, which can lead to a more calibrated output confidence. 
Instruction-tuning narrow the LLMs' exposure to specific tasks, reducing its ability to express uncertainty and making it more likely to provide overconfident outputs.}

\smallskip\noindent\textbf{\shin{3) Model size}}:
\shin{Large models consistently demonstrate improved reliability, as illustrated in Table~\ref{table:belief_icl_diff_sizes}.}


\subsection{\shin{Factors affecting the robustness}}
\label{sec:analysis_consistency}
\smallskip\noindent\textbf{\shin{1) Larger model cannot make zero-shot knowledge prompt more robust.}}
\shin{Similar to accuracy and reliability, few-shot knowledge prompts show improved robustness against accuracy fluctuations and consistency as model size increases. 
However, this effect is absent in zero-shot settings. 
For instance, the SD for the Llama2 family are 0.2014, 0.2131, and 0.2126 for the 7B, 13B, and 70B models, respectively. 
Similar inconsistencies are observed across other LLM families with varying model sizes. 
Refer Table~\ref{table:experiment-partial} for more details.}

\begin{figure}[t]
    \centering
    \includegraphics[width=\linewidth]{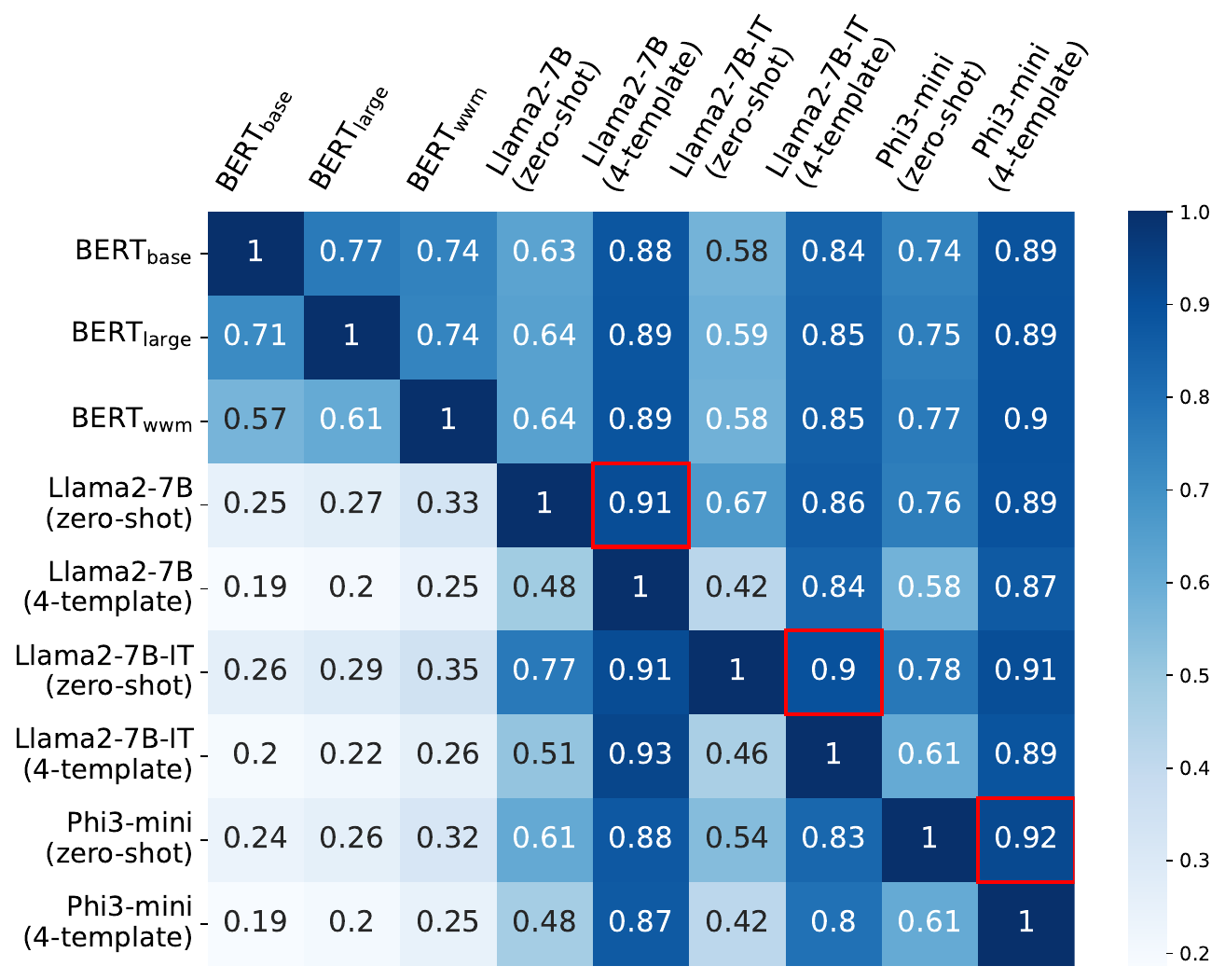}
    \caption{The knowledge sharing rate between models and ICL settings. Each cell indicates the rate of facts correctly predicted by the top-listed model relative to those captured by the left-listed model.}
    \label{fig:knowledge-coverage}
\end{figure}

\smallskip\noindent\textbf{\shin{2) Instruction-tuning making fluctuation less influenced by context.}}
\shin{Table~\ref{tab:pretrained_vs_instruction_tuned} and Table~\ref{table:experiment-full} show that the instruction-tuned models exhibit reduced fluctuation (smaller range and SD) in the zero-shot, 4-random, and 4-relation ICL settings, but perform worse in the 4-template setting. 
This suggests that instruction-tuned models become less influenced by context and more reliant on the instruction itself.}

\shin{In contrast, the Consist measure consistently decreased in instruction-tuned models, suggesting that while instruction-tuning improves instruction interpretation, it may weaken semantic understanding, especially with paraphrases.}

\subsection{How do PLMs perceive facts differently?}

Finally, we measure the differences in fact coverage among models. 
We first collect the correctly predicted facts for each template, defining these as the model's covered facts. 
Given the covered facts of two models, we measure the knowledge sharing rates using an asymmetric metric. This metric calculates the proportion of shared facts relative to each model's total covered facts.

Figure~\ref{fig:knowledge-coverage} shows the results. The average sharing rate among BERT models is 69.1\%, while it is 68.7\% for Llama2-7B, 
\shin{Llama2-7B-IT, and Phi3-mini in the zero-shot setting.} 
In comparison, the average sharing rate between encoder- and decoder-based PLMs reduced to 47.1\%. 
Meanwhile, knowledge sharing rates in both zero-shot and 4-template settings indicate that incorporating  examples improves the knowledge elicited by the PLMs. 
However, about 10\% of knowledge can still only be elicited in the zero-shot setting \shin{(see red boxes)}.


\begin{table}[t]
\small
\centering
\tabcolsep 4.3pt
    \begin{tabular}{lccc}
    \toprule
    \textbf{PLMs} & \textbf{Average} & \textbf{Maximum} & \textbf{Oracle} \\
    \midrule
     BERT$_\mathrm{wwm}$        & .1364   & .4501 & .6636 \\    \midrule

     Llama2-7B (zero-shot)                 & .3385   & .6577 & .8153 \\
     Llama3-8B (zero-shot)                 & .3427   & .7099 & \textbf{.8756} \\
     Phi3-small (zero-shot)                 & .4258   & .6828 & .8642 \\    \midrule

     Llama2-7B (4-template)                 & .6616 & .7197 &.8133  \\
     Llama3-8B (4-template)                 & \textbf{.7268} & \textbf{.7731} & .8628 \\
     Phi3-small (4-template)                 & .6612 & .7181 & .8346 \\
     \bottomrule
    \end{tabular}
    \caption{\textbf{Average}: mean accuracy using all templates; \textbf{Maximum}: accuracy with the best template for each relation; \textbf{Oracle}: accuracy when the best template is selected for each \yn{factual knowledge}.}
    \label{table:knowledge-coverage}
\end{table}

\section{Limitation of Prompt-based Probing}
\label{sec:upper-limit}

Finally, we examine the limitation of prompt-based knowledge probing \yn{by using our massively diverse dataset.}
First, we gauge the average knowledge coverage rate by using the average Acc@1 (\textbf{average}). 
Next, for each relation, we calculate the maximum Acc@1 using the template that yields the highest accuracy,\footnote{We select the prompt with the best subject expression among prompts for each fact.}
and use this value to estimate the upper limit of prompt-based knowledge probing (\textbf{maximum}).
Finally, we approximate the upper limit of facts contained in LLMs, by considering a fact as existing if at least one prompt of this fact can produce the correct answer (\textbf{oracle}).

\yn{Table~\ref{table:knowledge-coverage} shows three knowledge coverage rates on some PLMs.
For PLMs with zero-shot settings (including BERT$_\textbf{wwm}$), we observe nearly a 30\% increase between \textbf{average} and \textbf{maximum} accuracy, emphasizing the importance of selecting suitable templates for specific facts and the potential gains from prompt engineering. 
This gap can be reduced to 5\% with few-shot settings.}
\yn{However, the gap between \textbf{maximum} and \textbf{oracle} accuracy mostly remains.} This indicates that different facts prefer different templates, suggesting no \yn{versatile} template works for all facts. 
Combining templates reveals the true upper limits of PLMs' knowledge memorization and highlights the importance of using diverse prompts over optimizing a single one for retrieval.
Refer to \S\ref{sec:full-result-knowledge-coverage} for results on more PLMs.


\section{Related Work}

The LAMA probe was first proposed to evaluate the utility of \yn{PLMs as knowledge bases via solving the fill-in-the-blank task}~\cite{petroni-etal-2019-language}.
Several researchers extend the LAMA probe to evaluate PLMs' ability to understand facts from diverse linguistic aspects, such as the effect of negation/mispriming~\cite{kassner-schutze-2020-negated}, distractors~\cite{pandia-ettinger-2021-sorting}, multilingual understanding~\cite{keleg-magdy-2023-dlama, zhao-etal-2024-tracing} and models' consistency facing prompts with minor nuances~\cite{fierro-sogaard-2022-factual, elazar-etal-2021-measuring}.
However, these studies lack the inspection of PLMs' reliability in knowledge prediction, which is vital in deploying LLMs to real-world tasks. Moreover, solving the fill-in-the-blank task by LLMs with the causal LM objective can underestimate their knowledge recall ability.

\yn{Recently, QA-based datasets have been developed to evaluate the knowledge recall ability of decoder-only LMs. \citet{kamel2022} created a high-quality QA prompt set, which is further extended by \citet{wiland-etal-2024-bear} to evaluate
both causal and masked LMs. \citet{mallen-etal-2023-trust} and \citet{maekawa-etal-2024-retrieval} developed QA datasets to see the impact of knowledge popularity and retrieval augmentation. 
Since the writing style of these datasets is limited to questions, we cannot perform reliable robustness evaluation.}

\section{Conclusions}

\shin{This paper presents the multi-faceted factual probing benchmarks, BELIEF and BELIEF-ICL, for encoder- and decoder-based PLMs, respectively.}
Leveraging a multi-prompt dataset, BELIEFs provide various evaluation metrics, including accuracy, consistency, and reliability, enabling a thorough evaluation of PLMs' knowledge recall abilities.
To make BELIEFs more reliable, we build a new multi-prompt dataset for knowledge probing, MyriadLAMA, featuring diverse prompts for each fact.
We conducted extensive experiments of multiple encoder-based PLMs and recent LLMs. 

Based on the evaluation results, we identify key factors affecting the accuracy, reliability and robustness of PLMs' fact recall, such as model size, pre-training strategy and corpora, and ICL settings.
\shin{We also reveal the negative effect of instruction-tuning in recall factual knowledge from LLMs. 
This highlights the need for careful design of instruction-tuning to preserve LLMs' knowledge recall abilities.}
Finally, by probing facts in different ways, we find that PLMs hold more knowledge than what is revealed by using the optimal template, highlighting the limitations of prompt-based factual probing.

\section{Limitations}
MyriadLAMA contains an extensive amount of prompts, which leads to high evaluation costs. 
In the future, we aim to extract a diverse yet robust subset from MyriadLAMA to enable a more efficient evaluation of factual knowledge. 
Additionally, MyriadLAMA is built upon LAMA-UHN, which includes only 41 relationships. 
Expanding the range of relations is essential to improve coverage in the evaluation of factual knowledge. 
Lastly, we need to evaluate closed-source LLMs, such as GPT-4 and Claude, to examine performance differences between them and open-source LLMs.

\section*{Acknowledgements}
This work was partially supported by the special fund of Institute of Industrial Science, The University of Tokyo, by JSPS KAKENHI Grant Number JP21H03494, and by JST, CREST Grant Number JPMJCR19A4, Japan.

\bibliography{acl_latex}

\appendix
\section{Construction of MyriadLAMA}
\label{sec:dataset_creation}

In this appendix, we explain the detailed procedure for generating the derived triples from unique triples in MyriadLAMA.
As discussed in \S\ref{sec:MyriadLAMA}, this study first extends the unique triples contained in LAMA-UHN~\cite{Petroni2020HowCA} by searching new objects from T-REx~\cite{elazar-etal-2021-measuring}.
Next, for the obtained unique triples, we generate derived triples by combining concrete linguistic expressions associated with entities (``subjects'' and objects) and diversify relational templates using both manual labor and LLMs. We describe the detailed procedure as following.

\subsection{The extension of entities}

\paragraph{Extension of unique triples from T-REx}
LAMA-UHN is a refined subset derived from the LAMA dataset, which LAMA originates from T-REx~\cite{elsahar-etal-2018-rex}. T-REx is a large-scale knowledge base containing 11 million real-world knowledge triples, aligned with 3.09 million Wikipedia abstracts, designed to create large-scale alignments between Wikipedia abstracts and Wikidata triples. To achieve this alignment, T-REx employed three distinct aligners—NoSub, AllEnt, and SPO—each offering varying levels of accuracy (0.98, 0.96, and 0.88, respectively) as measured on a test set.
Despite the high alignment accuracy of all three aligners, LAMA-UHN selects only the triples aligned by NoSub, the aligner with the highest accuracy. While this choice ensures the high correctness of triples within LAMA, it potentially compromises the ability to fairly assess a PLM's knowledge recall ability, as it may overlook valid answers during evaluation.
To address this limitation, we expand the MyriadLAMA dataset by incorporating triples aligned by all three aligners—NoSub, AllEnt, and SPO—found in T-REx, based on the subject-relation pairs present in LAMA-UHN.
As the result, we increase the number of unique triples from 27,106 to 34,048 as shown in Table~\ref{tab:sample1}.

\paragraph{Extension of entities using aliases}
Next, we utilize aliases of entities obtained from Wikidata to acquire diverse linguistic expressions (and their paraphrases) for the ``subjects'' and objects.
Specifically, we used the Wikidata identifiers of entities\footnote{\url{https://www.wikidata.org/wiki/Wikidata:Identifiers}} and the Wikidata API\footnote{\url{https://www.wikidata.org/wiki/Special:EntityData/<entity_identifier>.json}} to retrieve the (English) alias expressions of entities. By combining the aliases of ``subjects'' and objects with the relation templates mentioned later, we generate numerous new derived triples. 
If $N$ ``subjects'' and $M$ objects are given for an unique triple, the number of derived triples according to this unique triple generated from a single relational template is $N \times M$.

\subsection{Diversification of relation templates}
We use a two-step procedure to create new relational  templates, to enhance ensure both the quality and quantity. 
Initially, we manually rewrite relational templates, ensuring that every relation has five templates. 
Then, we employ the generative LLM (GPT4) to automatically paraphrase 19 additional templates. In total, we produce 100 templates for each relation.

\paragraph{Step 1: Manually rewriting relational templates.}
The manual rewriting of the relational templates is performed by the first author of this paper.
We create new templates by describing the relationship between subject and object from different perspectives rather than creating templates with absolutely the same meaning with original template.  
Utilizing the resource provided by Wikidata \footnote{\url{https://www.wikidata.org/wiki/Property:<relation_identifier>}}, we not only paraphrase existing templates to generate new ones with diverse lexicons but also devise entailment expressions to encompass various semantic expressions that convey the same relations.
These newly created templates are guaranteed to uphold relational equivalence, following the relationship between the subject and object.
Taking P20 ([X] died in [Y].)\footnote{\url{https://www.wikidata.org/wiki/Property:P20}} as an example, we create new templates by either changing the sentence pattern or adding type information of object (e.g, [X] resided in [Y] until death). Furthermore, we also create templates without directly using the keywords of the relation (dead/death) but in a entailment way (\textit{e.g.}, [X] spent the last years of life in [Y].)
Moreover, we devise a question-answer style template for each relation to enhance syntactic diversity. In this template, the question incorporates the subject and relation information, while the answer corresponds to the object.

Note that, during the paraphrase, we observe that some templates in LAMA-UHN only partially express the original meaning of relations defined in Wikidata. These are inappropriate for specific knowledge triples.
For example, P136 describes the creative work's genre or an artist's field of work\footnote{\url{https://www.wikidata.org/wiki/Property:P136}}, which the type of work includes music, film, literature, etc. 
However, the original templates of P136 in LAMA-UHN is ``[X] plays [Y] music.,'' which cannot correctly retrieve information on work other than music.
For this kinds of template, we abandon the original templates and newly create five templates. 

\paragraph{Step 2: Paraphrasing templates using GPT-4}
Based on the original relation templates and the relation templates rewritten manually, we further paraphras these relation templates automatically using the GPT4-API (gpt-4-1106-preview\footnote{\url{https://platform.openai.com/docs/models/gpt-4-and-gpt-4-turbo}}) provided by OpenAPI. 
The instruction for paraphrasing used for GPT-4 generation is:
\begin{quote}
\textit{You are a professional tool that can paraphrase sentences into natural sentences that can correctly represent the relationship between [X] and [Y], without repetition. Make the paraphrase as diverse as possible using simple words. Please paraphrase the given sentence 19 times.}
\end{quote}
When the duplicated sentence is generated, we remove the duplication and regenerate new templates with the same instruction, until 19 different templates is generated. 
Furthermore, we observe that GPT-4 occasionally generates relation templates that are semantically inappropriate for specific relationships due to incorrect category information of entities. Consequently, in such instances, we refine the instructions to include the category information of the entities, ensuring accurate representation of the relationship between the subjects and the objects.
For example, when paraphrasing the relational template ``[X] used to work in [Y].''\footnote{\url{https://www.wikidata.org/wiki/Property:P937}}, we additionally add explicit guidance regarding the expected format and semantics of the relation templates to the above instruction, as following.
\begin{quote}
\textit{Be aware that [Y] is the geographic location but NOT company or organization, where persons or organizations were actively participating in employment, business or other work.}
\end{quote}
As a result, we can obtain the following paraphrased relational templates for ``[X] used to work in [Y].'':
\begin{itemize}
  \item ``[X] was formerly employed in [Y].''
  \item ``[X] once worked at [Y].''
  \item ``[Y] was the place where [X] used to be engaged in work.''
\end{itemize}

\subsection{Example of extended relational templates in MyriadLAMA} 
\label{sec:template_examples}

We display part of the created templates in MyriadLAMA. 
We randomly select two manually rewritten templates and three auto-generated templates of these two templates. 
We show the sampled templates for all the relations in Table\yn{s}~\ref{tab:template_example_1},~\ref{tab:template_example_2},~\ref{tab:template_example_3},~\ref{tab:template_example_4},~\ref{tab:template_example_5} and~\ref{tab:template_example_6}.

\section{The Advangage of MyriadLAMA}
\label{sec:dataset_comparision}

Given that our study seeks to mitigate the influence of individual prompt bias in evaluations, the availability of a wide range of prompts characterized in both quantity and diversity is crucial. 
The diversity ensures that different prompts can capture different aspects of the true knowledge distribution.
On the other side, the quality or correctness of prompts ensure the evaluation can accurately reflect the true knowledge recall ability. 

In this section, we provide a quantitative analysis of the quality and diversity of multi-prompt factual knowledge probing datasets.
The comparison results demonstrate the superiority of MyriadLAMA over previous datasets, enabling more accurate and comprehensive evaluations.
We conduct comparison between MyriadLAMA and other multi-prompts probing datasets, LPAQA~\cite{jiang-etal-2020-know} and PARAREL~\cite{elazar-etal-2021-measuring}, from the perspective of quantity and diversity. 

\subsection{Diversity comparison} 
We measure the diversity of multi-prompt factual knowledge probing datasets from both \textbf{quantity} and \textbf{linguistic diversity}. 

Specifically, we calculate the average number of prompts for each subject-relation pair as the \textbf{quantity} measure. 
MyriadLAMA introduces diversity into prompts by using various subject expressions and relational templates.
On average, MyriadLAMA provides 2.47 expressions for each subject.
In addition, we measure the \textbf{linguistic diversity} of relational templates from three aspects, as shown below:

\begin{description}
  \item[Lexicon:] We utilize the Jaccard distance of words in templates to gauge lexicon diversity. 
  \item[Syntax:] We adopt the syntax distance measure proposed in \cite{oya-2020-syntactic}, which calculates the distance between dependency trees. 
  \item[Semantics:] We quantify semantic diversity by calculating the L2 distance of sentence embeddings given by BERT$_\mathrm{large}$.
\end{description}

The results are shown in Table~\ref{table:dataset_comparison}. 
MyriadLAMA demonstrates superior quantity and diversity compared to existing multi-prompt factual probing datasets.
Although LPAQA exhibits greater semantic diversity, this is mainly due to its use of distant supervision to discover new templates.
This method often results in problematic templates that inadequately describe the relationships between subjects and objects. 
For example, for relation P937\footnote{\url{https://www.wikidata.org/wiki/Property:P937}} (``[X] used to work in [Y].''), LPAQA includes templates like ``[X] to meet [Y],'' which significantly deviate from the original semantic meaning. 
We analyze and compare template quality in the next section.

\begin{table}[t]
\small
\centering
\tabcolsep 3.3pt
    \begin{tabular}{lcccc}
    \toprule
    \multirow{3}{*}[3pt]{\textbf{Dataset}} & 
    \multirow{3}{*}[3pt]{\textbf{Quantity$\uparrow$}} & 
    \multicolumn{3}{c}{\parbox{2.3cm}{\centering\vspace{\fill}\textbf{Diversity$\uparrow$}\vspace{\fill}}} \\ 
    \cmidrule(lr){3-5} & 
     & \textbf{Lexicon} & \textbf{Syntax} & \textbf{Semantic}\\
    \midrule     
     PARAREL        & 7.30   & .4860 & .1489 & 11.03 \\ 
     LPAQA          & 53.27  & .5449 & .1713 & \textbf{13.55} \\     
     MyriadLAMA    & \textbf{263.47}   & \textbf{.6652} & \textbf{.2138} & 12.69\\\bottomrule
    \end{tabular}
    \caption{Comparison between multi-prompts datasets.}
    \label{table:dataset_comparison}
\end{table}

\subsection{Quality comparison}

In this section, we evaluate the quality of relational templates created MyriadLAMA in correctly expressing the relation between the subject and object. 
We manually evaluate the quality templates created in each dataset through a strict quality evaluation framework. 
Specifically, we evaluate each template based on its fluency and its ability to correctly express the semantic relationship between subjects and objects. 
Given the complex and specific constraints defined by Wikidata relations, creating perfect templates that satisfy all subjects and objects for a given relation is challenging.

\subsubsection{Semantic relationship between template and relation}
\smallskip\noindent\textbf{[Template $\subseteq$ Relation]: }
If the subject and object fit the template, it is correct for the relation, but the relation's knowledge range is broader than the template can cover. 
We denote such templates as \textbf{[Template $\subseteq$ Relation]}.
Using the templates in LAMA-UHN, which are often considered golden templates as examples, relation P136\footnote{\url{https://www.wikidata.org/wiki/Property:P1303}} uses the template ``[X] plays [Y] music.'' to describe creative work genres or an artist's field. 
However, P136 encompasses film, literature, and other arts, not just music. 

\smallskip\noindent\textbf{[Relation $\subseteq$ Template]: }
In contrast, If the subject-object pair is true for the relation, it is also true for the template, meaning the template's knowledge range is broader than the relation. 
For example, LAMA-UHN creates the template ``[X] died in [Y].'' for P20\footnote{\url{https://www.wikidata.org/wiki/Property:P20}}.
While this template can be used to infer a person's place of death, ``[Y]'' could also be the year ``[X]'' passed away.

\smallskip\noindent\textbf{[Relation $\cap$ Template > 0]: }
Additionally, some templates do not fit neatly into either \textbf{[Relation $\subseteq$ Template]} or \textbf{[Template $\subseteq$ Relation]} but can still correctly describe the relationship for some subject-object pairs.
For example, PARAREL, which paraphrases templates by human effort, uses ``[X] is a follower of [Y].'' to describe relation P140\footnote{\url{https://www.wikidata.org/wiki/Property:P140}} (religion or worldview). 
This template is appropriate for individuals but not for organizations, and it does not fully capture the relation's scope.
Therefore, it does not fit \textbf{[Relation $\subseteq$ Template]}.
Additionally, when ``[X]'' is a person, ``[Y]'' can be either a religious figure, leader, or the religion itself, which does not satisfy \textbf{[Template $\subseteq$ Relation]}. 
However, many subject-relation pairs can still be correctly captured by this template. We use \textbf{[Relation $\cap$ Template > 0]} to denote them.

\smallskip\noindent\textbf{[Irrelevant]: }
We consider templates that do not correctly convey the relationship between subject and object as \textbf{[Irrelevant]}. 
For example, LPAQA mined many irrelevant templates from the corpus without careful checking, resulting in low-quality templates, such as ``[X] arrived in [Y]'' for P937.

\smallskip\noindent\textbf{[Fully matching]: }
Finally, we use \textbf{[Fully matching]} to denote templates that can accurately capture all the subject-object pairs fitting in the relation.

\begin{figure}[t]
    \centering
    \includegraphics[width=\linewidth]{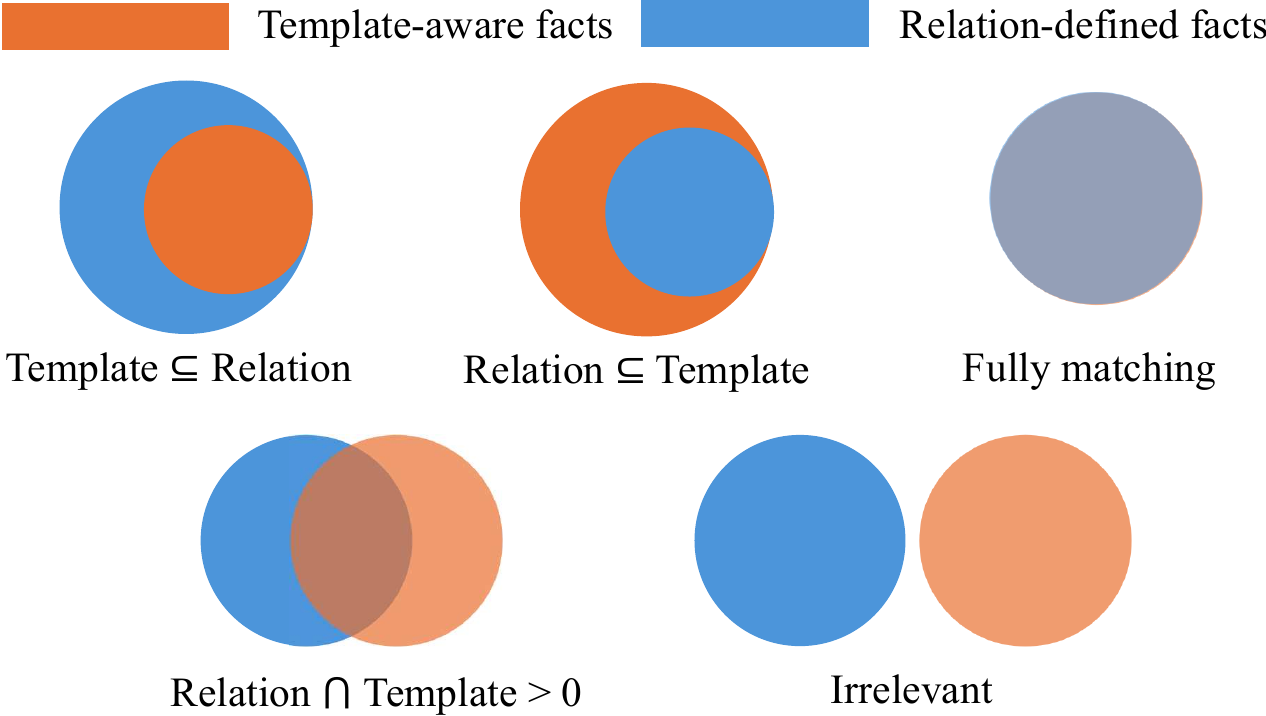}
    \caption{The relationship between the factual knowledge that can be covered by relation and template.}
    \label{fig:rel-vs-temp}
\end{figure}

We demonstrate the five types of semantic relationships between template and relation in Figure~\ref{fig:rel-vs-temp} using the Venn diagram. 

\subsubsection{Template quality evaluation metrics}\label{subsec:tempeval}
To accurately capture the fluency and the ability of created templates to correctly express the relationship between subjects and objects, we use the following metrics to score template quality. 
Each item is scored as either 1 or 0 based on whether the template meets the requirement.

\smallskip\noindent\textbf{1) Fluency:}
Is the template a natural sentence or noun phrase\footnote{We take the noun phrase into consideration as LPAQA created lots of noun phrase templates, such as ``[X], who works as [Y].`` for relation P106 (\url{https://www.wikidata.org/wiki/Property:P106})}? Set 1 if the templates are natural; Otherwise 0. 

\smallskip\noindent\textbf{2) \textbf{[Relation $\subseteq$ Template]}:} If the template can satisfy the definition of \textbf{[Template $\subseteq$ Relation]}, then 1; Otherwise 0.

\smallskip\noindent\textbf{3) \textbf{[Template $\subseteq$ Relation]}:} If the template can satisfy the definition of \textbf{[Template $\subseteq$ Relation]}, then 1; Otherwise 0.

\smallskip\noindent\textbf{4) \textbf{[Relation $\cap$ Template > 0]}:} If the template can satisfy the definition of \textbf{[Relation $\cap$ Template > 0]}, then 1; Otherwise 0.

If either \textbf{[Relation $\subseteq$ Template]} or \textbf{[Template $\subseteq$ Relation]} is 1, then \textbf{[Relation $\cap$ Template > 0]} must also be 1. 
If \textbf{[Relation $\subseteq$ Template]}, \textbf{[Template $\subseteq$ Relation]}, and \textbf{[Relation $\cap$ Template > 0]} are all 0, the template can is classified as \textbf{[Irrelevant]}. 
If all three metrics are 1, the template is classified as \textbf{[Fully Matching]}.

Specifically, as what PLMs see is the prompt with subject, we will consider the existence of subject when scoring the template. 
For example, P413\footnote{\url{https://www.wikidata.org/wiki/Property:P413}} describe the position or specialism of a player on a team. 
While the template ``[X] plays in the position of [Y].'' can be too general, as it could also describe a player's position in an orchestra, specifying ``[X]'' in the prompt reduces ambiguity, making it an accurate \textbf{[Fully Matching]} template to describe the relation.

\begin{table*}[ht]
\small
\centering
\tabcolsep 5.5pt
    \begin{tabular}{lccccc}
    \toprule
    \textbf{Dataset} & \textbf{Fluency} & \textbf{Template $\subseteq$ Relation} & \textbf{Relation $\subseteq$ Template} & \textbf{Relation $\cap$ Template > 0} & \textbf{Total Average}\\
    \midrule     
    LAMA-UHN       & 1    & .732 & .976 & 1     & 3.707 \\ 
     PARAREL       & 0.99 & .790 & .905 & .985 & 3.670 \\ 
     LPAQA         & 0.57 & .220 & .345 & .405 & 1.540 \\     
     MyriadLAMA    & 1    & .770 & .830 & .985 & 3.585 \\\bottomrule
    \end{tabular}
    \caption{The average quality metrics across templates from various factual probing datasets. LAMA-UHN has 41 templates, all of which were used for evaluation. For the other three multi-prompt probing datasets, we sampled 200 templates for comparison.}
    \label{table:dataset_quality_comparison}
\end{table*}

\subsubsection{Evaluation result and analysis}
The comparison includes the 4 dataset: LAMA-UHN, PARAREL, LPAQA and MyriadLAMA. 
Considering the amount of all the templates in the three models (6654 templates in total), we randomly sample 200 templates for multi-prompt probing datasets and use all 41 templates in LAMA-UHN for evaluation. 
To ensure objectivity, we anonymize the source of each template and mix them together for annotator (the first author). 
We publicize the annotation results here\footnote{\url{https://anonymous.4open.science/r/belief-CC8A}}.
The evaluation result is shown in Table~\ref{table:dataset_quality_comparison}. 

From the Table~\ref{table:dataset_quality_comparison}, We observe that our semi-automatically generated relational templates achieve quality comparable to manually created datasets like LAMA-UHN and PARAREL, while being 100 times larger than LAMA-UHN and 13.7 times larger than LPAQA. MyriadLAMA significantly outperforms LPAQA in template quality due to our two-stage template creation method.

\begin{figure}[t]
    \centering
    \includegraphics[width=0.9\linewidth]{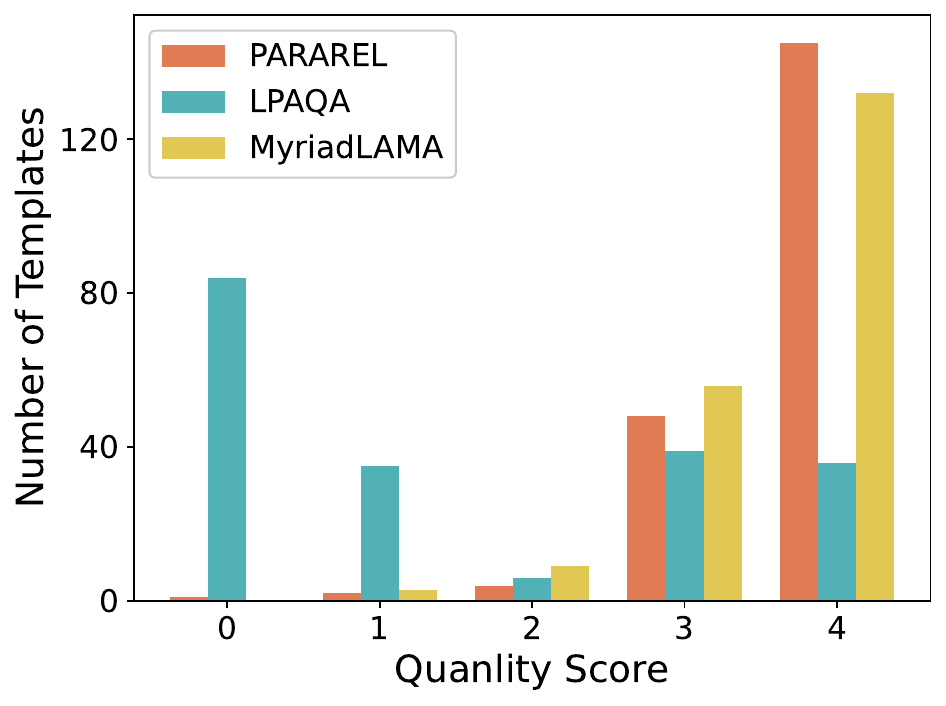}
    \caption{Quality score distributions over multi-prompt factual probing datasets.}
    \label{fig:quality-score-distribution}
\end{figure}

Furthermore, Figure~\ref{fig:quality-score-distribution} shows the score distributions for 200 templates across the three multi-prompt datasets. 
It reveals that LPAQA has many low-score templates, with 0 being the most common score.
Compared to PARAREL, MyriadLAMA has slightly more templates with a score of 3 but slightly fewer with a score of 4, resulting in slightly lower overall quality.

\section{Ablation Analysis of MyriadLAMA} 
\label{sec:dataset_evaluation}
In this section, we conduct ablation analysis MyriadLAMA to understand the validity of diversification on entities and templates.

\subsection{Validity of extended entity expressions}
We evaluate the validity of the extended entity expressions in MyriadLAMA by checking if these extensions cover facts that PLMs can capture but are missed in LAMA-UHN due to strict entity expression limitations. We conduct this analysis on BERT models, focusing on facts with extended subject and object expressions. MyriadLAMA contains 13,123 facts with extended subjects and 23,195 facts with extended objects. We measure the rate at which extended subjects/objects achieve higher ranks than the original expressions in the token distribution output.

The results, shown in the Table~\ref{table:entity_validity} below, indicate that around 50\% of extended subjects and 20\% of extended objects achieve higher ranks than the original entities. This suggests that many facts are missed in LAMA-UHN and other single-expression factual knowledge probing datasets.

\begin{table}[t]
\small
\centering
    \begin{tabular}{lcccc}
    \toprule
    \textbf{PLMs} & \textbf{Subject} & \textbf{Object} \\
    \midrule
     BERT$_\mathrm{base}$   & .5355 & .2107\\
     BERT$_\mathrm{large}$  & .5358 & .2116 \\
     BERT$_{\mathrm{wwm}}$  & .5272 & .1853 \\\bottomrule
    \end{tabular}
    \caption{Rate of prompts where the extend entities gain higher probability than the original entity.}
    \label{table:entity_validity}
\end{table}

\subsection{Validity of paraphrased templates}
In this section, we evaluate the validity of the relation templates in MyriadLAMA.
We investigate the accuracy of each template and compare the accuracies between templates in LAMA-UHN, manually rewritten templates and auto-generate templates. 
Specifically, for each relation, we evaluate the accuracy (Acc@1) of all relation templates separately, and then calculate the minimum, and maximum accuracies among all templates for each relation. 
We then measure the dataset-level minimum/maximum accuracy by micro-averaging the template set with the minimum/maximum template accuracies (41 templates in each set). 
Finally, all of the template-specific accuracies are then micro-averaged to compute the average Acc@1. 
As indicated in Table \ref{table:dataset_acc_eval}, while the quality of MyriadLAMA's prompts significantly varies, the high-quality prompts are notably superior to those of LAMA-UHN. 
Although the average accuracy of MyriadLAMA is lower than that of LAMA-UHN, it is considered that this is because MyriadLAMA uses relation templates that have been semi-automatically created, whereas LAMA-UHN uses carefully selected entities and templates.

\begin{table}[t]
\small
\centering
    \begin{tabular}{lcccc}
    \toprule
    \multirow{3}{*}[3pt]{\textbf{PLMs}} & 
    \multirow{3}{*}[2pt]{\textbf{LAMA-UHN}} & 
    \multicolumn{3}{c}{\textbf{MyriadLAMA}} \\ 
     \cmidrule(lr){3-5} & 
    \textbf{} & \textbf{Min} & \textbf{Max} & \textbf{Mean} \\
    \midrule
     BERT$_\mathrm{base}$      & .2403 & .0000 & .3534 & .1103 \\
     BERT$_\mathrm{large}$     & .2454 & .0007 & .3728 & .1185 \\
     BERT$_{\mathrm{wwm}}$     & .2448 & .0015 & .3695 & .1453 \\\bottomrule
    \end{tabular}
    \caption{Acc@1 of MyriadLAMA and LAMA-UHN}
    \label{table:dataset_acc_eval}
\end{table}

\begin{table}[t]
\small
\centering
\tabcolsep 3.2pt
    \begin{tabular}{lcccc}
    \toprule
    \multirow{3}{*}[3pt]{\textbf{PLMs}} & 
    \multicolumn{2}{c}{\textbf{Consist}$\uparrow$} & 
    \multicolumn{2}{c}{\parbox{2.3cm}{\centering\vspace{\fill}\textbf{Acc@1 range\\(min/max)}\vspace{\fill}}} \\ 
    \cmidrule(lr){2-3} \cmidrule(lr){4-5} & 
    \textbf{Subject} & \textbf{Relation} & \textbf{Subject} & \textbf{Relation} \\
    \midrule
     BERT$_\mathrm{base}$      & .5745 & .1504 & .0673/.1441 & .0000/.3534 \\
     BERT$_\mathrm{large}$     & .5497 & .1548 & .0714/.1554 & .0007/.3728 \\
     BERT$_{\mathrm{wwm}}$     & .5005 & .1057 & .0831/.1884 & .0015/.3695 \\ \bottomrule
    \end{tabular}
    \caption{Diversity evaluation of subjects and relation templates}
    \label{table:dataset_cons_eval}
\end{table}

\subsection{What matters to robustness? Diverse subject vs. templates}

Next, we aim to investigate the factors contributing to varying performance and inconsistent predictions of prompts. 
MyriadLAMA creates diverse prompts for each fact by combining different subject expressions and templates.
To gauge their impact on robustness, we examine both the consistency (Consist) and the accuracy range (min/max) across various expressions of subjects or relations, assessed individually.
To achieve this, the complete set of prompts was partitioned into multiple subsets, with each subset containing only one expression for each unique subjects or relations.
The Acc@1 of the prompts obtained in this manner is then evaluated using different variants of BERT.

The results in Table~\ref{table:dataset_cons_eval} indicate that while the accuracy range (min/max) and consistency (Consist) caused by aliases of subjects is less pronounced compared to diverse expressions of relational templates, its effect on factual knowledge evaluation remains significant. 
These findings highlight the vulnerability of factual knowledge evaluation based on single prompts and underscore the significance of harnessing the diversity of prompts within MyriadLAMA for robust assessments.

\subsection{Manually rewritten vs. auto-generated templates}
Upon comparing relational templates generated through manual rewriting and GPT-4 auto-generation, we find that auto-generated templates exhibit comparable quality (accuracy) to manually rewritten templates; they also demonstrate less diversity in acquiring different predictions, aligning with our expectations.

To assess the validity of LLM-generated templates for knowledge probing, we rank the accuracies (Acc@1) of manually created templates against those generated by LLMs. 
Specifically, for each relation, we rank the 5 manual templates among all 100 templates and calculate the average rank across all manually created templates for all relations. 
Table~\ref{table:prompts_compare} shows the average Acc@1 ranks of manual templates among 100 templates on BERT$_\mathrm{base}$, BERT$_\mathrm{large}$, BERT$_\mathrm{wwm}$. They are 47.40, 45.64, and 44.80, respectively. These values closely approximate the average rank of 50, indicating that auto-generated templates can achieve nearly the same accuracy as manually created templates. 

Furthermore, we quantify the diversity discrepancy between manually written and auto-generated templates. We categorize the auto-generated templates, including the original ones, as one group, resulting in five groups for each relation, each comprising 20 templates. 
Subsequently, we evaluate the similarity between templates within the same group and across different groups using the consistency measure (Consist), as presented in Table~\ref{table:prompts_compare}. 
The consistency among prompts within the same group (inner-group) is notably high, whereas prompts from different groups (inter-group) exhibit less diversity in predictions. 
This underscores the significance of manual phrase rewriting, which can yield more diverse prompts and facilitate a more comprehensive evaluation.

\begin{table}[t]
\small
\centering
\tabcolsep 2.8pt
    \begin{tabular}{lccc}
    \toprule
    \multirow{3}{*}[3pt]{\textbf{PLMs}} & 
    \multirow{3}{*}[3pt]{\parbox{2.3cm}{\centering\vspace{\fill}\textbf{Average rank of\\manual prompts\\based on Acc@1}\vspace{\fill}}} & 
    \multicolumn{2}{c}{\textbf{Consist}} \\ 
    \cmidrule(lr){3-4} & 
     & \textbf{Inner-group} & \textbf{Inter-group} \\
    \midrule
     BERT$_\mathrm{base}$      & 47.40 & .2904 & .1065 \\
     BERT$_\mathrm{large}$     & 45.64 & .2884 & .1125 \\
     BERT$_{\mathrm{wwm}}$     & 44.80 & .2387 & .0630 \\ \bottomrule
    \end{tabular}
    \caption{Comparison between prompts generated through manual labor and LLM.}
    \label{table:prompts_compare}
\end{table}

\section{QA-Style ICL and Its Evaluation}
\label{sec:qa-icl}

\subsection{QA-style instruction}
Beside the mask-prediction-style (MP-style) ICL task, we also defined and evaluate a question-answer-style (QA-style) ICL task utilizing the QA-style relational templates available in MyriadLAMA. 
This is available because MyriadLAMA provides 20 QA-style templates for each relation, offering not only syntactical diversity but also suitability for the autoregressive generation process in LLMs. Each QA-style prompt adheres to a format where the subject and relation construct the question, and the object corresponds to the answer, such as \texttt{``Who developed [X]? [Y].''}
For the QA prompt, we employ the few-shot prompt comprising $X$ random QA pairs, following the format outlined in InstructGPT~\cite{NEURIPS2022_b1efde53}. 
Given that all objects in MyriadLAMA are intended to be matched with single words, we append the instruction \texttt{``Answer each question in one word.''} to ensure compatibility. 

Given the limited number of templates (20 for each relation) in the QA-style, the evaluation of QA-style prompts represents only one-fifth of the full prompts in MyriadLAMA.

\subsection{Evaluation}
\label{sec:qa-icl-experiment}
We measure the ability of QA-style prompts in adhering to instructions and compare it to MP-style prompts. 
To ensure a fair comparison between QA- and MP-style ICL, we conduct evaluations using shared templates in both settings on Llama2-7B, with 20 QA-style templates for each relation.

We evaluate the abilities of fact prediction and one-word generation individually on Llama2-7B using average Acc@1 and rate of the one-word generation.
As demonstrated in Table~\ref{table:instruct_follow}, Llama2-7B exhibits a remarkable capability to comprehend instructions for answering questions and generating one-word answers.
We observe that QA-style instructions perform better under the zero-shot setting, likely due to decoder-based PLMs' ability to autoregressively generate text. 
However, this gap diminishes with the use of few-shot examples. This suggests that while MP-style prompts may slightly underestimate the knowledge in LLMs in zero-shot settings, MP-style ICL settings can demonstrate comparable or even superior performance in factual knowledge prediction compared to QA-style ICL prompts.

\begin{table}[t]
\small
\centering
    \begin{tabular}{lcccc}
    \toprule
    \multirow{3}{*}[3pt]{\textbf{ICL settings}} & 
    \multicolumn{2}{c}{\textbf{\makecell{Fact prediction\\(Acc@1)}}} & 
    \multicolumn{2}{c}{\parbox{2.3cm}{\centering\vspace{\fill}\textbf{1-word ratio}\vspace{\fill}}} \\ 
    \cmidrule(lr){2-3} \cmidrule(lr){4-5} & 
    \textbf{QA} & \textbf{MP} & \textbf{QA} & \textbf{MP} \\
    \midrule
     zero-shot      & .4534 & \textbf{.5066} & \textbf{.5285} & .4802 \\
     4-random       & .5429 & \textbf{.5591} & .7996 & \textbf{.8058} \\
     4-relation     & .6582 & \textbf{.6649} & .9187 & \textbf{.9246} \\
     4-template     & .6687 & \textbf{.6765} & .9216 & \textbf{.9266} \\
     \bottomrule
    \end{tabular}
    \caption{Instruction following rate on Llama2-7B.}
    \label{table:qa-vs-mp}
\end{table}

\section{Examples of BELEIF-ICL Prompts}
\label{sec:prompt_examples}

In this section, we provide example prompts for the four patterns introduced in \S\ref{sec:belief-icl}: zero-shot, X-random, X-relation, and X-template. We focus on examples where X equals 4, which is the primary setting used in our work.

\subsection{zero-shot}
\texttt{Predict the [MASK] in each sentence in one word.\\
Q: [MASK] consists of LAUPT.\\
A:}

\subsection{4-random}
\texttt{Predict the [MASK] in each sentence in one word.\\
Q: [MASK] is the administrative center of Jiangsu.\\
A: Nanjing.\\
Q: Mar del Plata and [MASK] are sister cities that have been developing together.\\
A: Havana.\\
Q: Malawi has established diplomatic ties with [MASK].\\
A: Australia.\\
Q: Which country is House of Representatives located? [MASK].\\
A: Libya.\\
Q: [MASK] consists of LAUPT.\\
A:
}

\subsection{4-relation}

\texttt{Predict the [MASK] in each sentence in one word.\\
Q: What is the overarching group for Panzer Division Kempf? [MASK].\\
A: Wehrmacht.\\
Q: To whom does Mount Bulusan relate? [MASK].\\
A: Luzon.\\
Q: Who is responsible for Army National Guard? [MASK].\\
A: National Guard.\\
Q: What group is pharmacy a part of? [MASK].\\
A: biology.\\
Q: [MASK] consists of environmental factors.\\
A:}

\subsection{4-template}
\texttt{Predict the [MASK] in each sentence in one word.\\
Q: [MASK] consists of Panzer Division Kempf.\\
A: Wehrmacht.\\
Q: [MASK] consists of Mount Bulusan.\\
A: Luzon.\\
Q: [MASK] consists of Army National Guard.\\
A: National Guard.\\
Q: [MASK] consists of pharmacy.\\
A: biology.\\
Q: [MASK] consists of environmental factors.\\
A:}

\section{Experimental Details}
\label{sec:experiment-details}

In this section, we list the detailed information of PLMs used in our study, including 3 encoder-based models and 8 decoder-based LLMs. 

\subsection{Model cards}
Here are the links from Hugging Face to load each model:
\begin{description}
    \item[\small BERT$_\mathrm{base}$:] 
    \small \url{https://huggingface.co/bert-base-uncased}
    \item[\small BERT$_\mathrm{large}$:] 
    \small \url{https://huggingface.co/bert-large-uncased}
    \item[\small BERT$_\mathrm{wwm}$:] 
    \small \url{https://huggingface.co/bert-large-uncased-whole-word-masking}
    \item[\small ALBERT$_\mathrm{base}$:] 
    \small \url{https://huggingface.co/albert/albert-base-v2}
    \item[\small ALBERT$_\mathrm{large}$:] 
    \small \url{https://huggingface.co/tftransformers/albert-xxlarge-v2}
    \item[\small Llama2-7B:] 
    \small \url{https://huggingface.co/meta-llama/Llama-2-7B} 
    \item[\small Llama2-7B-IT:] 
    \small \url{https://huggingface.co/meta-llama/Llama-2-7B-hf} 
    \item[\small Llama2-13B:] 
    \small \url{https://huggingface.co/meta-llama/Llama-2-13B} 
    \item[\small Llama2-13B-IT:] 
    \small \url{https://huggingface.co/meta-llama/Llama-2-13B-hf} 
    \item[\small Llama2-70B:] 
    \small \url{https://huggingface.co/meta-llama/Llama-2-70B} 
    \item[\small Llama2-70B-IT:] 
    \small \url{https://huggingface.co/meta-llama/Llama-2-70B-hf} 
    \item[\small Llama3-8B:] 
    \small \url{https://huggingface.co/meta-llama/Meta-Llama-3-8B}
    \item[\small Llama3-8B-IT:] 
    \small \url{https://huggingface.co/meta-llama/Meta-Llama-3-8B-Instruct}
    \item[\small Llama3-70B:] 
    \small \url{https://huggingface.co/meta-llama/Meta-Llama-3-70B}
    \item[\small Phi3-mini:] 
    \small \url{https://huggingface.co/microsoft/Phi-3-mini-4k-instruct}
    \item[\small Phi3-small:]
    \small \url{https://huggingface.co/microsoft/Phi-3-small-8k-instruct}
    \item[\small Phi3-medium:] 
    \small \url{https://huggingface.co/microsoft/Phi-3-medium-4k-instruct}

\end{description}

\begin{table*}[t]
\small
\centering
\setlength{\tabcolsep}{4.2pt} 
\begin{tabular}{llrrrl} 
    \toprule
    \multirow{2}{*}{\textbf{LLMs}} & 
    \multirow{2}{*}{\textbf{Architecture}} & 
    \multirow{2}{*}{\textbf{IT\dag}} & 
    \multirow{2}{*}{\textbf{Model size}} & 
    \multicolumn{2}{c}{\textbf{Pre-training corpora}} \\ 
    \cmidrule(lr){5-6}
    & & & & \textbf{Size} & \textbf{Resource} \\
    \midrule
    BERT$_\mathrm{base}$    & Encoder-based & No & 110M &  \multirow{5}{*}{$\left.\begin{array}{@{}r@{}}\text{3.3B words}\\ \text{3.3B words}\\ \text{3.3B words}\\ \text{3.3B words}\\ \text{3.3B words} \end{array}\right\rbrace$} & \multirow{5}{*}{\makecell[l]{
    BookCorpus (11,038 unpublished books) and \\English Wikipedia\\(excluding lists, tables, and headers)}} \\ 
    BERT$_\mathrm{large}$   & Encoder-based & No & 336M &  \\ 
    BERT$_\mathrm{wwm}$     & Encoder-based & No & 336M &   \\ 
    ALBERT$_\mathrm{base}$  & Encoder-based* & No & 11.8M &  \\ 
    ALBERT$_\mathrm{large}$ & Encoder-based* & No & 223M & \\ 
    \midrule
    Llama2-7B      & Decoder-based & No & 7B  & \multirow{6}{*}{$\left.\begin{array}{@{}r@{}}\text{2.0T tokens}\\ \text{2.0T tokens}\\ \text{2.0T tokens}\\ \text{2.0T tokens}\\ \text{2.0T tokens}\\ \text{2.0T tokens} \end{array}\right\rbrace$} & \multirow{6}{*}{\makecell[l]{Publicly available online data \\(excluding sites containing personal info;\\ factual knowledge sources are upsampled)}} \\ 
    Llama2-13B     & Decoder-based & No & 13B & \\ 
    Llama2-70B     & Decoder-based & No & 70B &  \\ 
    Llama2-7B-IT   & Decoder-based & Yes& 13B & \\ 
    Llama2-13B-IT  & Decoder-based & Yes& 13B &  \\ 
    Llama2-70B-IT  & Decoder-based & Yes& 70B &  \\ 
    \midrule
    Llama3-8B      & Decoder-based & No & 8B  & \multirow{3}{*}[1.2pt]{$\left.\begin{array}{@{}r@{}}\text{15T+ tokens}\\ \text{15T+ tokens}\\ \text{15T+ tokens} \end{array}\right\rbrace$} & \multirow{3}{*}{\makecell[l]{Publicly available online data \\(details unknown, code is 4x larger than Llama2)}} \\ 
    Llama3-8B-IT   & Decoder-based & No & 8B &  \\ 
    Llama3-70B     & Decoder-based & No & 70B &  \\ 
    \midrule
    Phi3-mini      & Decoder-based & Yes & 3.8B & \multirow{3}{*}[1.2pt]{$\left.\begin{array}{@{}r@{}}\text{4.9T tokens}\\ \text{4.9T tokens}\\ \text{4.9T tokens} \end{array}\right\rbrace$} & \multirow{3}{*}{\makecell[l]{High-quality materials including educational data,\\ textbook-like generated text, high-quality chats}} \\ 
    Phi3-small     & Decoder-based & Yes & 7B  &  \\ 
    Phi3-medium    & Decoder-based & Yes & 14B &  \\ 
    \bottomrule
    \multicolumn{6}{l}{\small \dag Specify whether the model is an instruction-tuned model or not} \\
    \multicolumn{6}{l}{\small *ALBERT shares parameter between token embeddings and transformer layers to compress parameters.} \\
\end{tabular}
\caption{The pre-training information of various PLMs used in this study}
\label{table:llm_info}
\end{table*}

\subsection{Model differences}
We outline the differences between PLMs in their pre-training details in Table~\ref{table:llm_info}, including the type of Transformer architecture, model size, and the size and resources of the pre-training corpora.

\subsection{Evaluation results on all PLMs based on BELIEFs}
\label{sec:full-result}
We present all evaluation results and their computational costs in this section. In Table~\ref{table:experiment-full}, we report the full-scale experiments using all the prompts provided by MyriadLAMA. This includes PLMs with 8B parameters or fewer, such as BERT$_\mathrm{base}$, BERT$_\mathrm{large}$, BERT$_\mathrm{wwm}$, ALBERT$_\mathrm{base}$, ALBERT$_\mathrm{large}$, Llama2-7B, Llama3-8B, Llama2-7B-IT, Llama3-8B-IT, Phi3-mini (3.8B), and Phi3-small (7B). 
For decoder-based models, we conduct experiments on 4 types of ICL settings.

For PLMs with more than 8B parameters, we report the evaluation results using partial prompts from MyriadLAMA, specifically using manually-rewritten templates (5 per relation), which account for 1/20 of the prompts compared to the full-scale experiments.
Meanwhile, we run these models on only two ICL settings: zero-shot and 4-template. 
To ensure fair comparison with models having 8B parameters or fewer, we apply the same settings to all other decoder-based LLMs.
The result is shown in Table~\ref{table:experiment-partial}.
We also list the approximate runtime for each experiments in these two tables.
The experiments are all run on the NVIDIA RTX 6000 Ada. 
For experiments using model less than or equal to 8B parameters, we use single GPU to measure the consumption time. 
We use 2 GPUs for Llama2-13B and Phi3-small and 4 GPUs for 70B models. 

\begin{figure*}[htbp]
    \centering
    \subcaptionbox{BERT and its variants}[0.3\textwidth]{\includegraphics[width=0.3\textwidth]{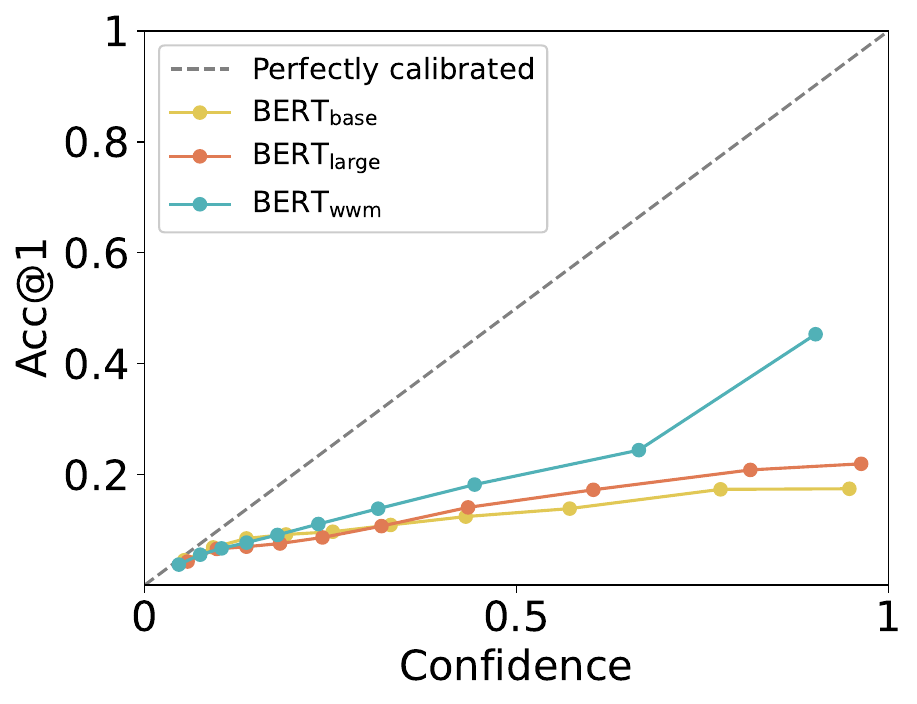}}
    \subcaptionbox{ALBERT and its variants}[0.3\textwidth]{\includegraphics[width=0.3\textwidth]{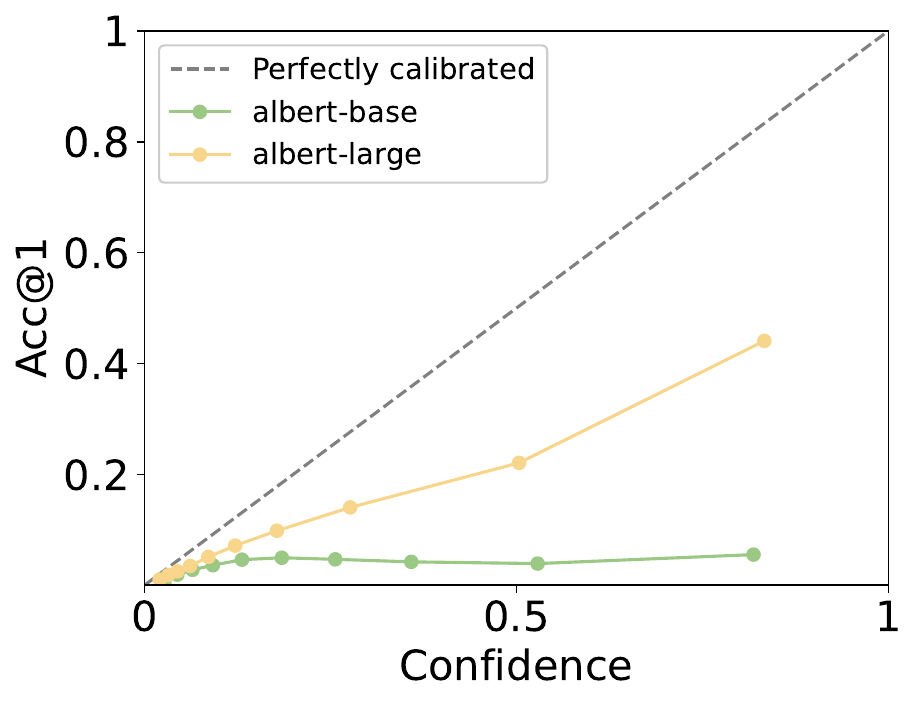}}
    \subcaptionbox{Llama2-7B with 4 ICL settings}[0.3\textwidth]{\includegraphics[width=0.3\textwidth]{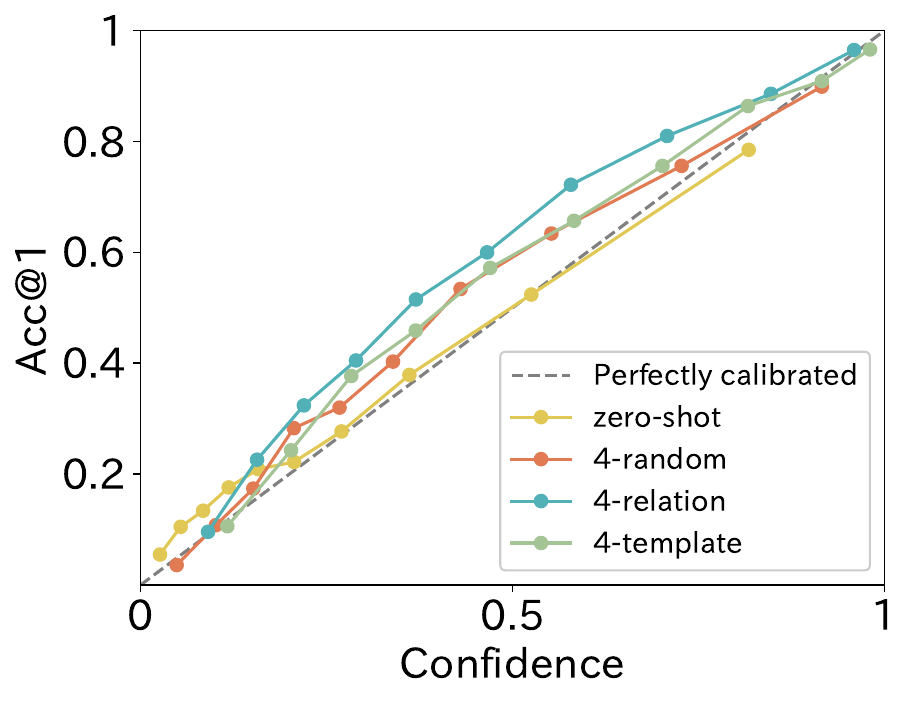}}
    \subcaptionbox{Llama2-7B-IT with 4 ICL settings}[0.3\textwidth]{\includegraphics[width=0.3\textwidth]{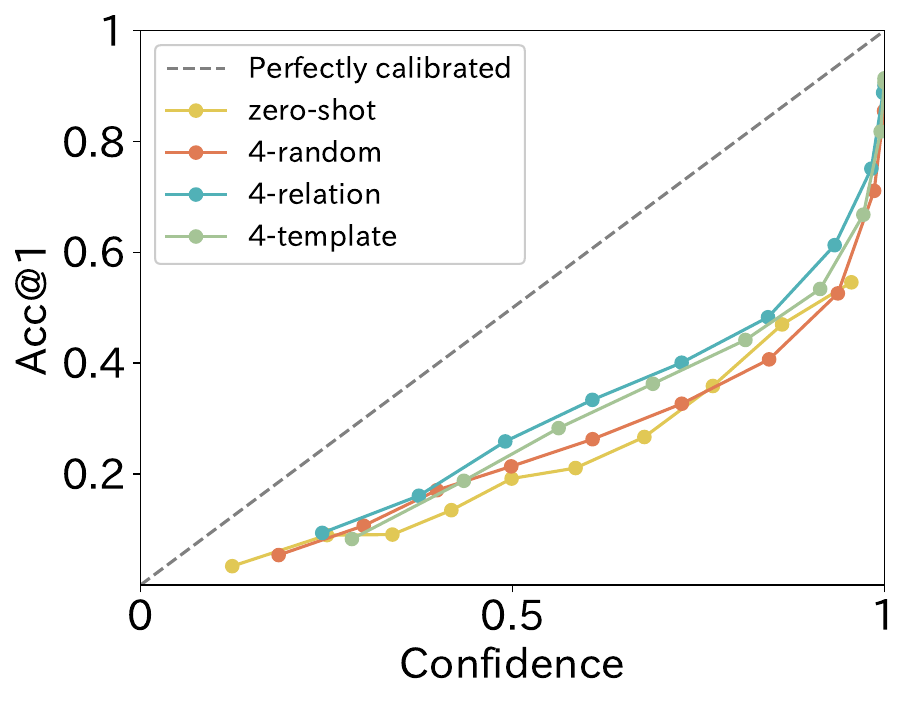}}
    \subcaptionbox{Llama3-8B with 4 ICL settings}[0.3\textwidth]{\includegraphics[width=0.3\textwidth]{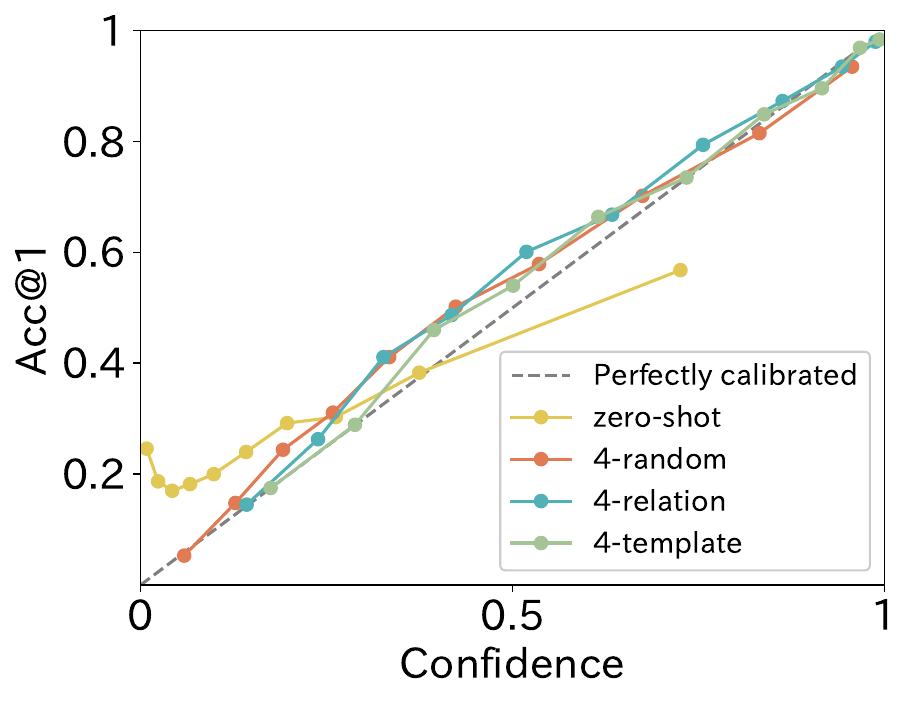}}
    \subcaptionbox{Llama3-8B-IT with 4 ICL settings}[0.3\textwidth]{\includegraphics[width=0.3\textwidth]{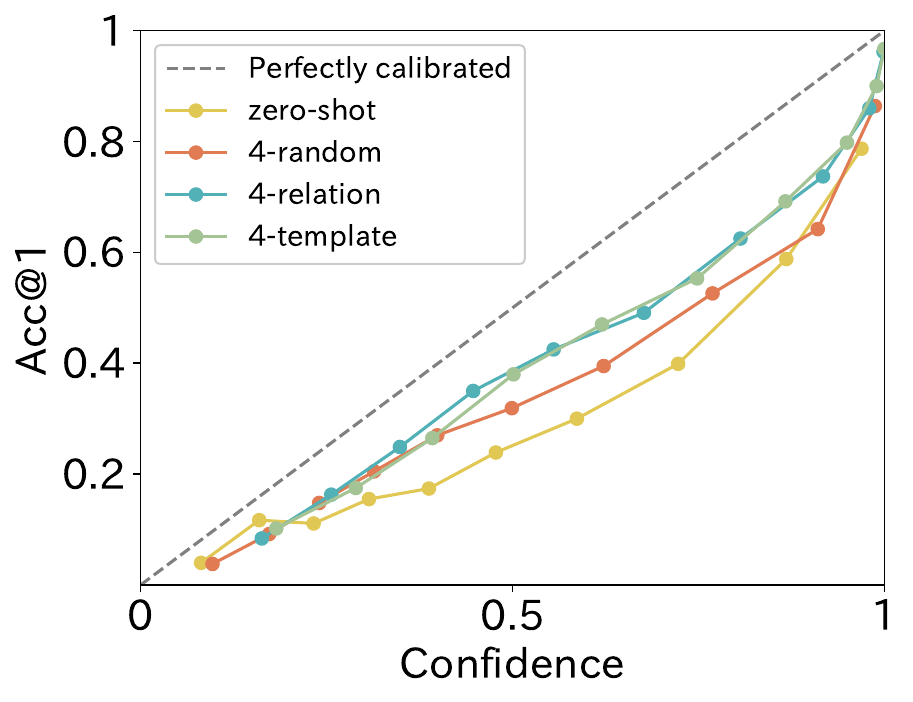}}
    \subcaptionbox{Phi3-mini with 4 ICL settings}[0.3\textwidth]{\includegraphics[width=0.3\textwidth]{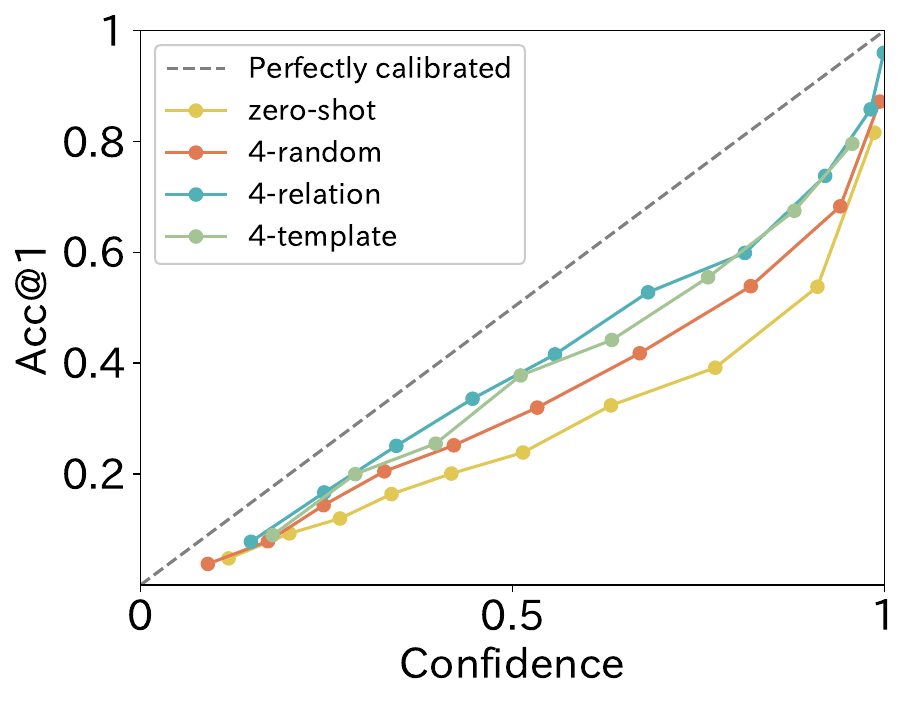}}
    \subcaptionbox{Phi3-small with 4 ICL settings}[0.3\textwidth]{\includegraphics[width=0.3\textwidth]{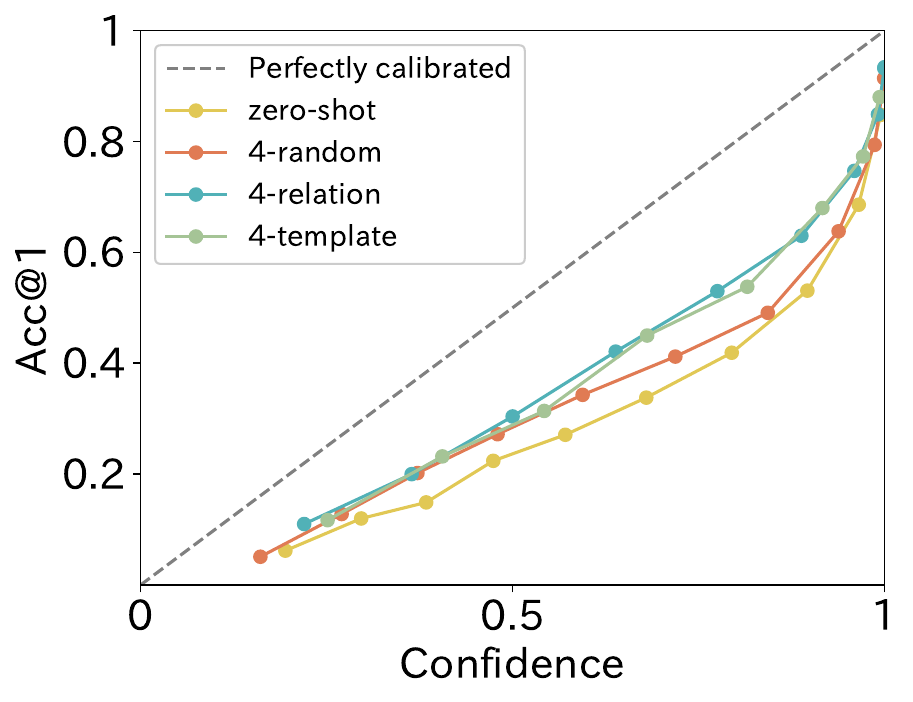}}
    \caption{The calibration between accuracy and confidence of PLMs' prediction with full-scale experiments.}
    \label{fig:overconf-full}
\end{figure*}

\begin{figure*}[htbp]
    \centering
    \subcaptionbox{Llama2-7B with 4 ICL settings}[0.3\textwidth]{\includegraphics[width=0.3\textwidth]{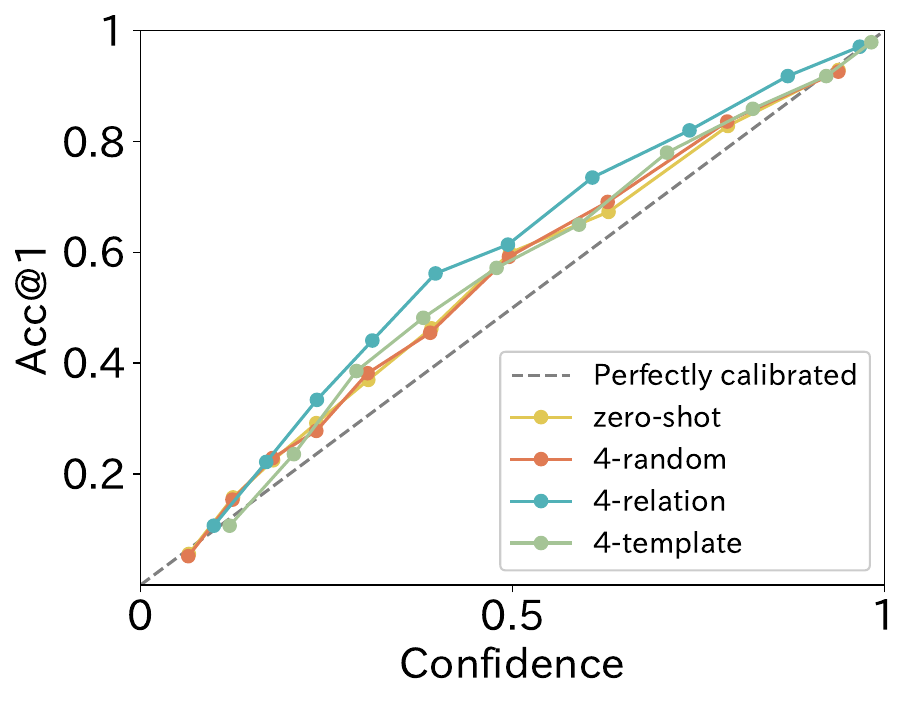}}
    \subcaptionbox{Llama2-7B-IT with 4 ICL settings}[0.3\textwidth]{\includegraphics[width=0.3\textwidth]{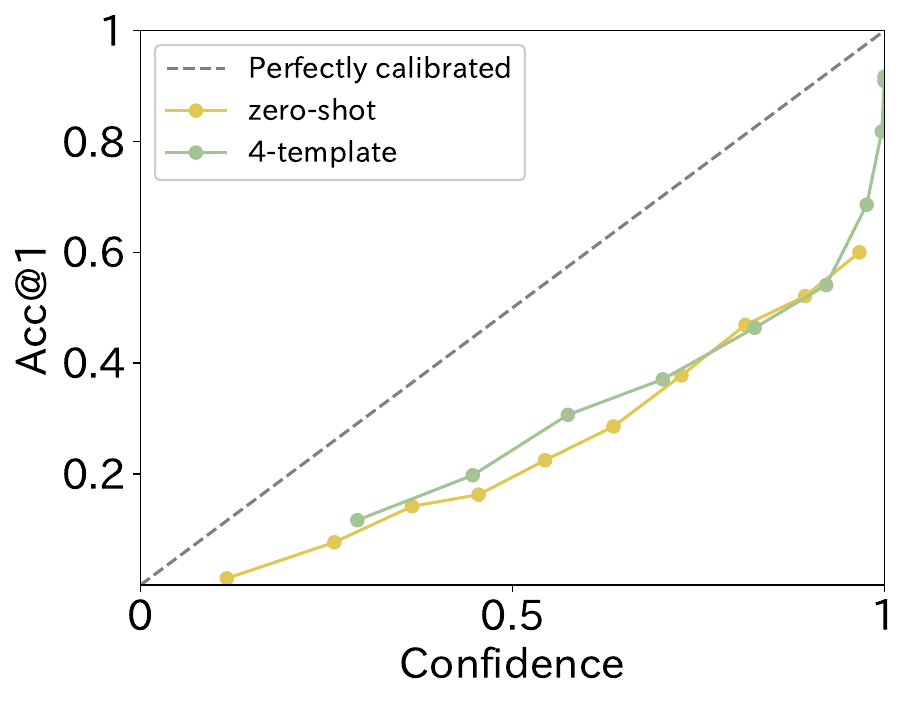}}
    \subcaptionbox{Llama2-13B with 2 ICL settings}[0.3\textwidth]{\includegraphics[width=0.3\textwidth]{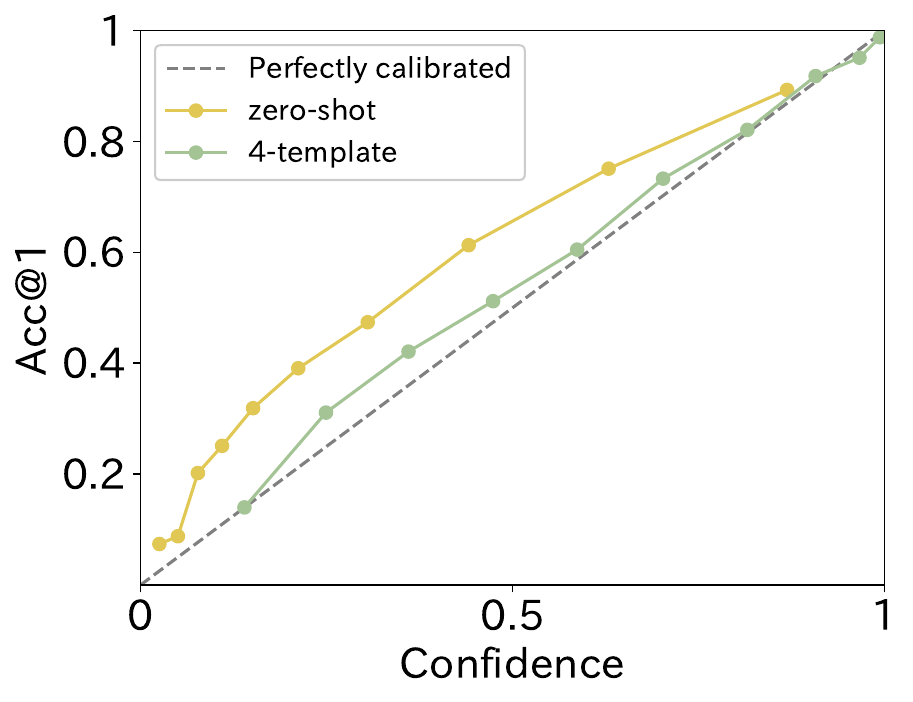}}
    \subcaptionbox{Llama2-13B-IT with 2 ICL settings}[0.3\textwidth]{\includegraphics[width=0.3\textwidth]{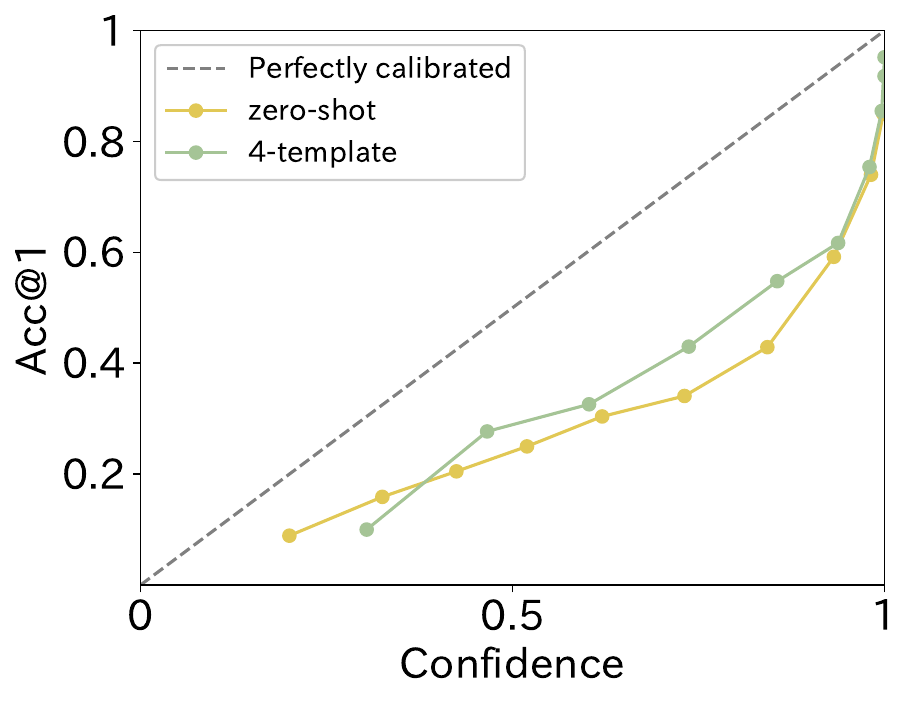}}
    \subcaptionbox{Llama2-70B with 2 ICL settings}[0.3\textwidth]{\includegraphics[width=0.3\textwidth]{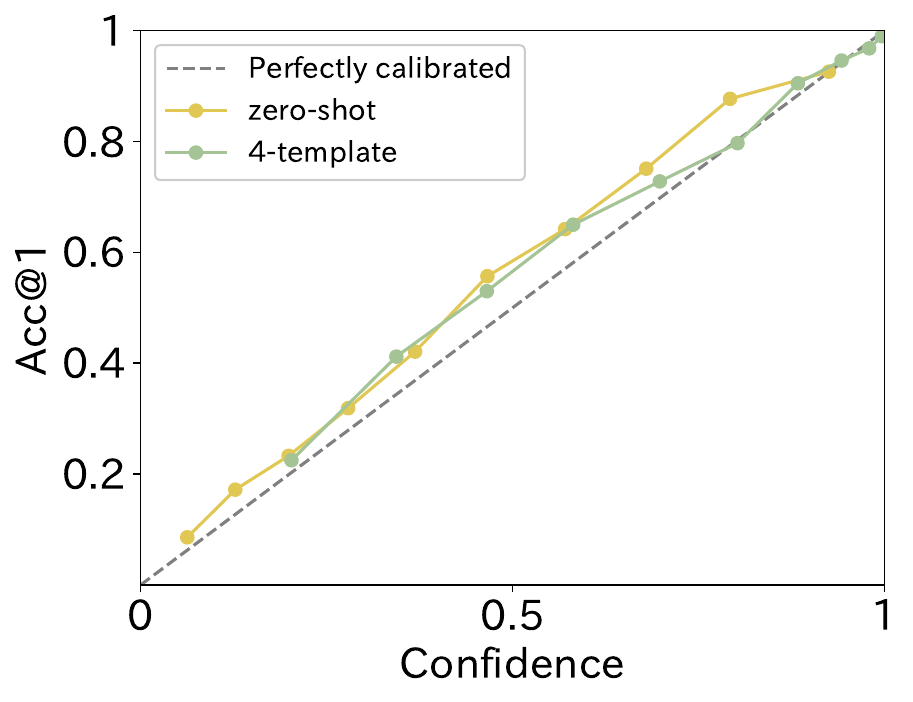}}
    \subcaptionbox{Llama2-70B-it with 2 ICL settings}[0.3\textwidth]{\includegraphics[width=0.3\textwidth]{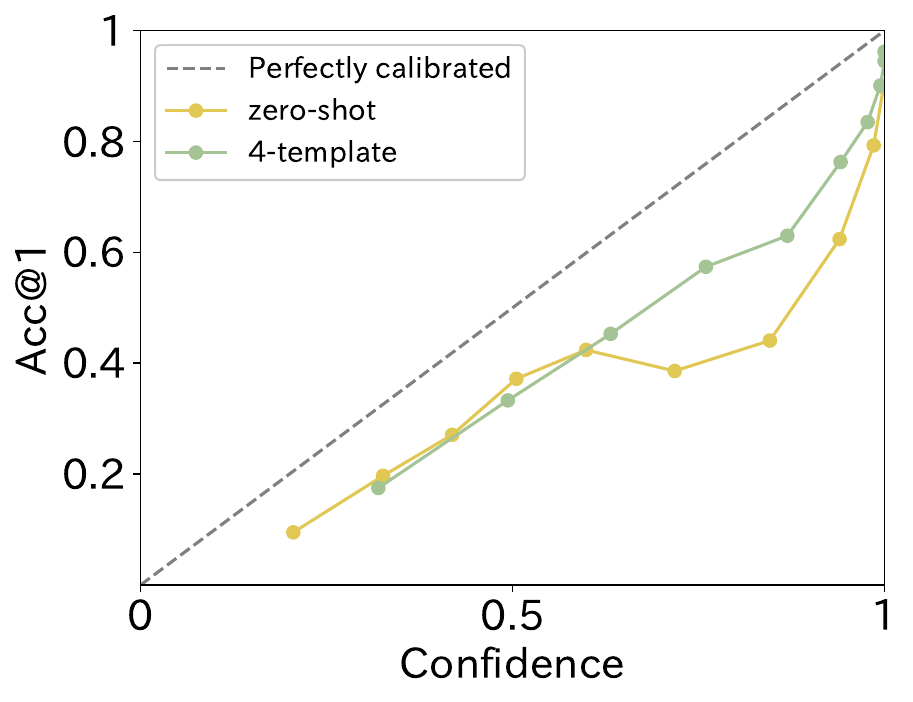}}
    \subcaptionbox{Llama3-8B with 2 ICL settings}[0.3\textwidth]{\includegraphics[width=0.3\textwidth]{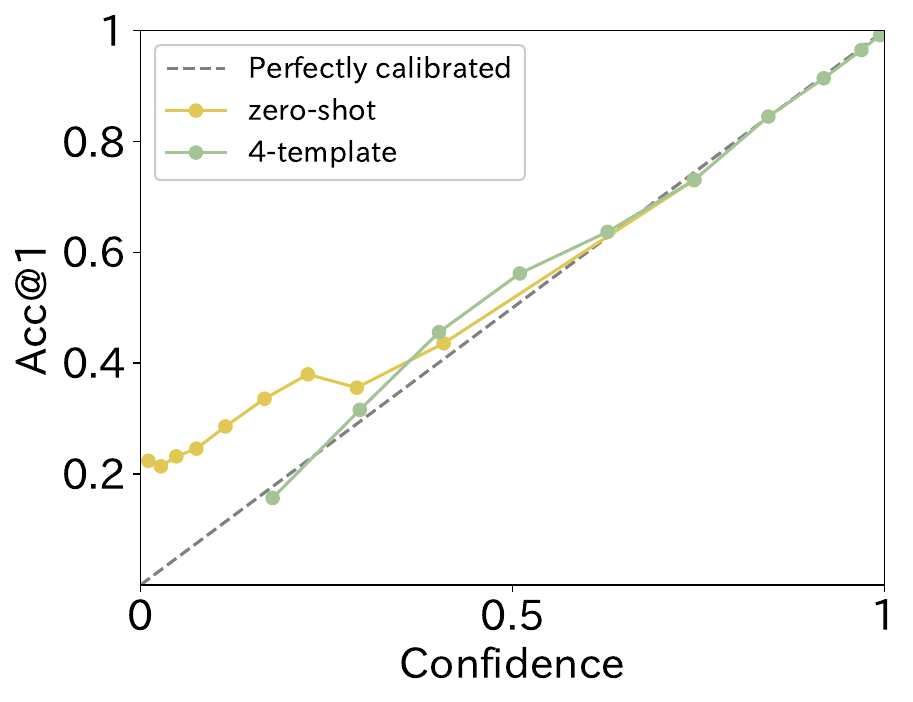}}
    \subcaptionbox{Llama3-8B-IT with 2 ICL settings}[0.3\textwidth]{\includegraphics[width=0.3\textwidth]{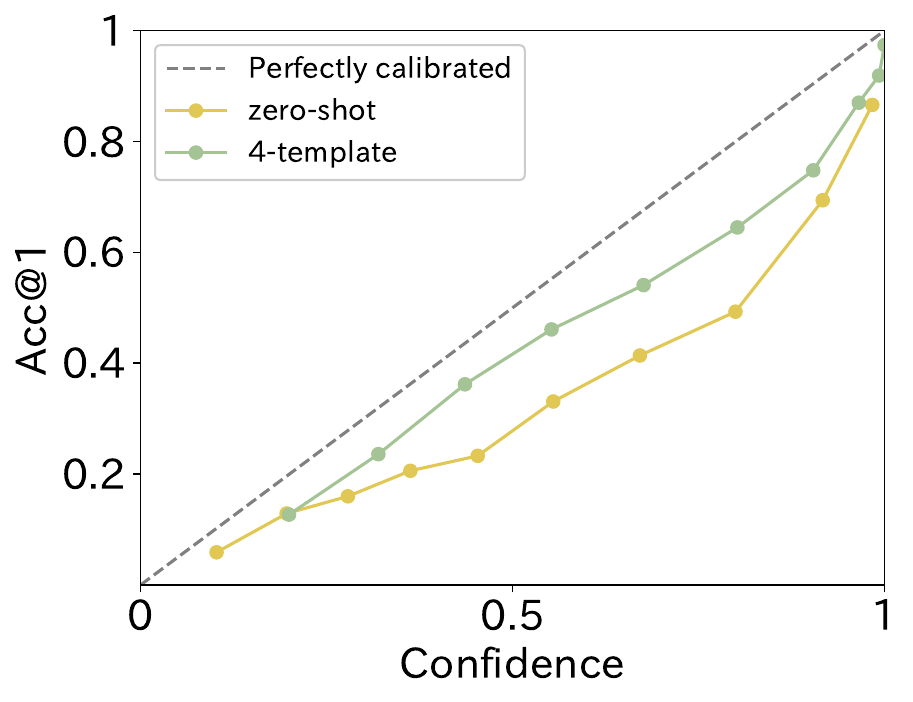}}
    \subcaptionbox{Llama3-70B with 2 ICL settings}[0.3\textwidth]{\includegraphics[width=0.3\textwidth]{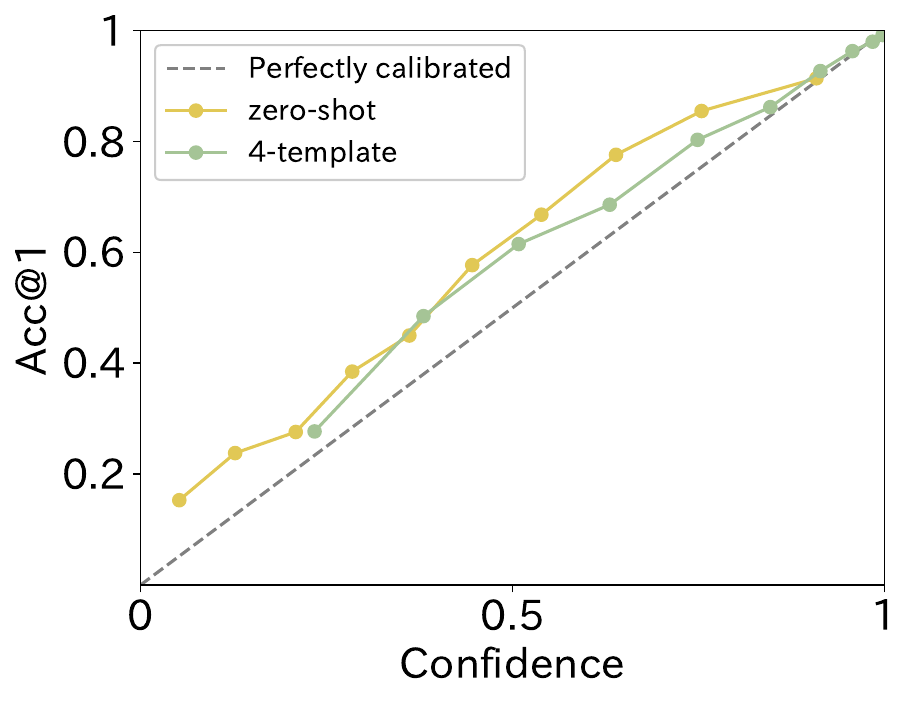}}
    \subcaptionbox{Phi3-mini with 4 ICL settings}[0.3\textwidth]
    {\includegraphics[width=0.3\textwidth]{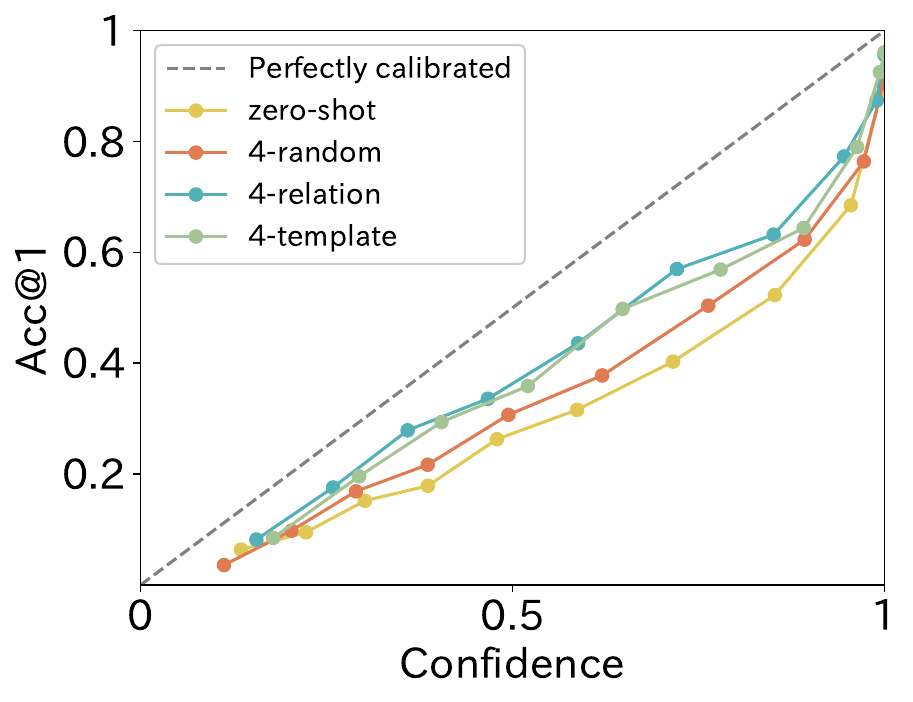}}
    \subcaptionbox{Phi3-small with 4 ICL settings}[0.3\textwidth]{\includegraphics[width=0.3\textwidth]{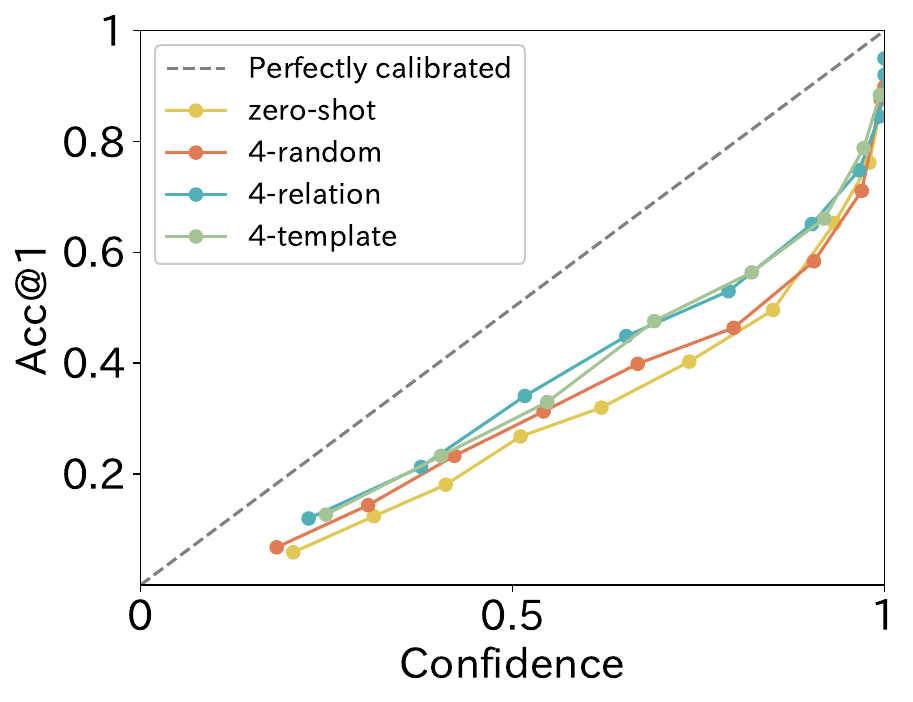}}
    \subcaptionbox{Phi3-medium with 2 ICL settings}[0.3\textwidth]{\includegraphics[width=0.3\textwidth]{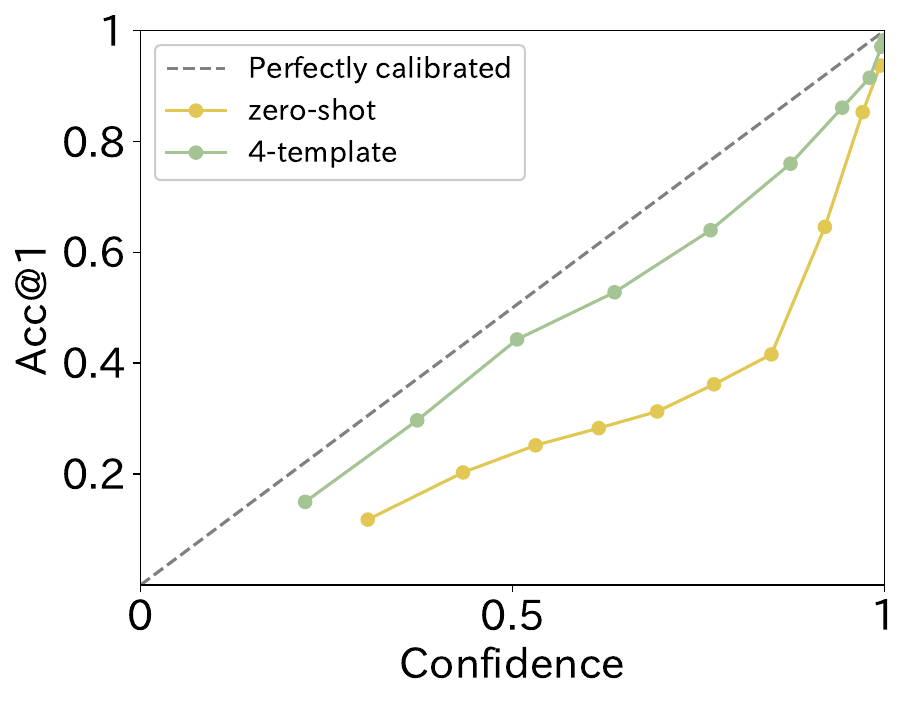}}
    \caption{The calibration between accuracy and confidence of PLMs' prediction with partial-scale experiments (using manually rewritten templates only).}
    \label{fig:overconf-partial}
\end{figure*}

Furthermore, we display the calibration level between accuracy and confidence for a straightforward inspection on Ovconf metrics. 
We show the calibration figures of models with full-scale experiments in Figure~\ref{fig:overconf-full} and experiments with partial prompts in Figure~\ref{fig:overconf-partial}.

\begin{table*}[ht]
\small
\centering
    \begin{tabular}{llccccccr}
    \toprule
    \multicolumn{2}{c}{\multirow{2}{*}[-2pt]{\textbf{PLMs}}} & 
    \multirow{2}{*}[-2pt]{\textbf{Acc@1}} & 
    \multicolumn{2}{c}{\textbf{Fluctuation$\downarrow$}} &
    \multirow{2}{*}[-2pt]{\textbf{Consist$\uparrow$}} & 
    \multirow{2}{*}[-2pt]{\textbf{Ovconf}} & 
    \multirow{2}{*}[-2pt]{\textbf{1-word ratio$\uparrow$}} & 
    \multirow{2}{*}[-2pt]{\textbf{Runtime}} \\
        \cmidrule(lr){4-5} & & &\textbf{range} & \textbf{SD} \\
    
    \midrule
    \multirow{3}{*}{BERT} &
     BERT$_\mathrm{base}$      & .1095 & .1534 & .0217 & .1682 & .2154 & N/A & 6.3h\\
     & BERT$_\mathrm{large}$   & .1102 & .1574 & .0220 & \textbf{.1713} & .2052 & N/A & 7.4h\\
     & BERT$_\mathrm{wwm}$     & \textbf{.1364} & \textbf{.1517} & \textbf{.0208} & .1524 & \textbf{.1000} & N/A & 7.4h\\
     \midrule
    \multirow{2}{*}{ALBERT} &
     ALBERT$_\mathrm{base}$      & .0362 & \textbf{.0668} & \textbf{.0131} & \textbf{.1333} & .1647 & N/A & 6.1h\\
     & ALBERT$_\mathrm{large}$   & \textbf{.0974} & .1110 & .0148 & .0821 & \textbf{.0553} & N/A & 15.2h\\ \midrule
    \multirow{4}{*}{Llama2-7B} &
     zero-shot      & .3385 & .2602 & .0299 & .1269 & -.1119 & .4752 & 46.4h\\
     & 4-random       & .4816 & .2250 & .0270 & .2312 & \textbf{-.0894} & .8247 & 47.8h\\
     & 4-relation     & .6286 & .1221& .0150 & .3753 & -.1335 & .9060  & 47.8h\\
     & 4-template     & \textbf{.6616} & \textbf{.0294} & \textbf{.0036} & \textbf{.4163} & -.0933 & \textbf{.9299}  & 47.8h\\
     \midrule
    \multirow{4}{*}{Llama2-7B-IT} &
     zero-shot      & .2925 & .1980 & .0253 & .1151 & .2605 & .9069 & 46.4h\\
     & 4-random       & .4334 & .1958 & .0229 & .2128 & .2410 & .9081 & 47.8h \\
     & 4-relation     & .5576 & .0791& .0092 & .3341 & \textbf{.1900} & .9314 & 47.8h \\
     & 4-template     & \textbf{.5896} & \textbf{.0439} & \textbf{.0050} & \textbf{.3687} & .2061 & \textbf{.9380} & 47.8h \\
     \midrule
     \multirow{4}{*}{Llama3-8B} &
     zero-shot      & .3427 & .2864 & .0350 & .0240 & -.1329 & .1572 & 44.9h\\
     & 4-random       & .5205 & .2033 & .0273 & .2156 & -.0796 & .8147 & 82.1h\\
     & 4-relation     & .6871 & .1236 & .0156 & .3659 & -.0783 & .9071 & 82.1h\\
     & 4-template     & \textbf{.7268} & \textbf{.0220} & \textbf{.0026} & \textbf{.4015} & \textbf{-.0582} & \textbf{.9187} & 82.1h\\
     \midrule
     \multirow{4}{*}{Llama3-8B-IT} &
     zero-shot      & .3578 & .2213 & .0262 & .1660 & .1402 & .7925 & 44.9h\\
     & 4-random       & .4290 & .2068 & .0222 & .2137 & .1038 & .8511 & 82.1h\\
     & 4-relation     & .5727 & .0731 & .0092 & .3239 & .0760 & .9140 & 82.1h\\
     & 4-template     & \textbf{.6508} & \textbf{.0372} & \textbf{.0040} & \textbf{.3727} & \textbf{.0800} & \textbf{.9331} & 82.1h\\
     \midrule
     \multirow{4}{*}{Phi3-mini (3.8B)} &
     zero-shot      & .3498 & .2374 & .0292 & .1465 & .1752 & .8641 & 30.7h\\
     & 4-random       & .4193 & .2324 & .0269 & .1649 & .1189 & .8184 & 32.9h\\
     & 4-relation     & .5686 & .1440 & .0164 & .2818 & \textbf{.0755} & .8769 & 32.9h\\
     & 4-template     & \textbf{.6067} & \textbf{.0510} & \textbf{.0048} & \textbf{.3612} & .0887 & \textbf{.8808} & 32.9h\\ \midrule
     \multirow{4}{*}{Phi3-small (7B)} &
     zero-shot      & .4258 & .2437 & .0292 & .1782 & .2171 & .8883 & 82.4h\\
     & 4-random       & .4889 & .2170 & .0276 & .2070 & .1670 & .8913 & 148h\\
     & 4-relation     & .6339 & .1012 & .0129 & .3361 & \textbf{.1252} & .9287 & 148h\\
     & 4-template     &  \textbf{.6612} & \textbf{.0360} & \textbf{.0043} & \textbf{.3626} & .1279 & \textbf{.9411} & 148h\\
    \bottomrule
    \end{tabular}
    \caption{Evaluation results and approximate experiment runtime on all MyriadLAMA data. Comparison of LLMs with 8B parameters or less, including BERT and its variants, across four ICL settings.}
    \label{table:experiment-full}
\end{table*}

\begin{table*}[ht]
\small
\centering
    \begin{tabular}{llccccrcr}
    \toprule
    \multicolumn{2}{c}{\multirow{2}{*}[-2pt]{\textbf{PLMs}}} & 
    \multirow{2}{*}[-2pt]{\textbf{Acc@1$\uparrow$}} & 
    \multicolumn{2}{c}{\textbf{Fluctuation$\downarrow$}} &
    \multirow{2}{*}[-2pt]{\textbf{Consist$\uparrow$}} & 
    \multirow{2}{*}[-2pt]{\textbf{Ovconf}} & 
    \multirow{2}{*}[-2pt]{\textbf{1-word ratio$\uparrow$}} & 
    \multirow{2}{*}[-2pt]{\textbf{Runtime}} \\
        \cmidrule(lr){4-5} & 
        \textbf{} & \textbf{} &\textbf{Range} & \textbf{Stedv} \\
    \midrule
    \multirow{12}{*}{zero-shot} &
     Phi3-mini (3.8B)   & .4248 & .1880 & .0247 & .2066 & .1609 & .8596 & 1.54h\\
     & Phi3-small (7B)  & .4881 & .1900 & .0244 & .2284 & .1985 & \textbf{.8996} & 4.12h\\
     & Llama2-7B        & .4311 & .2014 & .0249 & .1932 & \textbf{-.0922} & .5558 & 2.32h\\
     & Llama2-7B-IT     & .3566 & .1862 & .0228 & .1932 & .2417  & .8961 & 2.32h\\
     & Llama3-8B        & .4224 & .2820 & .0353 & .1269 & -.1438 & .1786 & 2.45h\\
     & Llama3-8B-IT     & .4279 & .1962 & .0217 & .2337 & .1260  & .9179 & 2.45h\\
     & Llama2-13B       & .4785 & .2131 & .0260 & .1437 & -.1673 & .3185 & 4.84h\\
     & Llama2-13B-IT    & .4639 & \textbf{.1701} & \textbf{.0222} & .2358 & .2180  & .7542 & 4.84h\\
     & Phi3-medium (14B)& .5173 & .2123 & .0277 & \textbf{.6167} & .2316  & .7759 & 4.85h\\
     & Llama2-70B       & .5675 & .2126 & .0280 & .2598 & -.0988 & .6239 & 28.97h\\
     & Llama2-70B-IT    & .5223 & .2055 & .0259 & .2489 & .1608  & .7891 & 28.97h\\
     & Llama3-70B       & \textbf{.5974} & .2137 & .0278 & .2290 & -.1438 & .7790 & 32.55h\\
     \midrule
     \multirow{12}{*}{4-template} &
     Phi3-mini (3.8B)   & .6106 & .0314 & .0039 & .3686 & .0911  & .9051 & 1.65h\\
     & Phi3-small (7B)  & .6668 & .0306 & .0039 & .3666 & .1222  & .9413 & 7.40h\\
     & Llama2-7B        & .6699 & .0257 & .0034 & .4174 & -.0933 & .9299 & 2.39h\\
     & Llama2-7B-IT     & .6013 & .0368 & .0045 & .3629 & .2007  & .9372 & 2.39h\\
     & Llama3-8B        & .7316 & .0194 & .0025 & .4060 & -.1119 & .9190 & 4.10h\\
     & Llama3-8B-IT     & .6563 & .0252 & .0032 & .3752 & .0535  & .9315 & 4.10h\\
     & Llama2-13B       & .7080 & .0235 & .0031 & .4326 & -.0662 & .9190 & 4.23h\\
     & Llama2-13B-IT    & .6482 & .0301 & .0038 & .3656 & .1708  & .9341 & 4.23h\\
     & Phi3-medium (14B)& .7304 & .0207 & .0025 & .4009 & \textbf{.0317} & .9350 & 3.88h\\
     & Llama2-70B       & .7784 & .0190 & .0024 & .4449 & -.0690 & .9256 & 21.99h\\
     & Llama2-70B-IT    & .7232 & .0258 & .0031 & .4226 & .1026  & \textbf{.9582} & 21.99h\\
     & Llama3-70B       & \textbf{.8211} & \textbf{.0139} & \textbf{.0017} & \textbf{.4636} & -.0812 & .9378 & 43.10h\\
     \bottomrule
    \end{tabular}
    \caption{Evaluation results and approximate experiment runtime on partial MyriadLAMA data. Comparison of all LLMs (8 models) using zero-shot and 4-template ICL settings with manually rewritten templates.}
    \label{table:experiment-partial}
\end{table*}

\subsection{Knowledge coverage rate on all PLMs}
\label{sec:full-result-knowledge-coverage}
We present the average, maximum, and upper limit knowledge coverage rates, as introduced in \S\ref{sec:upper-limit}, for all PLMs evaluated using all templates.
The results are shown in Figure~\ref{table:full-scale-knowledge-coverage}.

\begin{table*}[ht]
\small
\centering
    \begin{tabular}{llccc}
    \toprule
    \multicolumn{2}{c}{\textbf{PLMs}} & \textbf{Average} & \textbf{Maximum} & \textbf{Upper Limit} \\
    \midrule
    \multirow{3}{*}{BERT}
    & BERT$_\mathrm{base}$       & .1095   & .4248 & .6209 \\
    & BERT$_\mathrm{large}$      & .1102   & .4451 & .6556 \\
    & BERT$_\mathrm{wwm}$        & .1364   & .4501 & .6636 \\ \midrule
    \multirow{2}{*}{ALBERT}
    & ALBERT$_\mathrm{base}$     & .0362   & .2175 & .3405 \\
    & ALBERT$_\mathrm{large}$    & .0974   & .3746 & .5979 \\ \midrule
    \multirow{4}{*}{Llama2-7B}
    & zero-shot                  & .3385   & .6577 & .8153 \\
    & 4-random                   & .4816   & .7026 & .8587 \\
    & 4-relation                 & .6286   & .7179 & .8475 \\
    & 4-template                 & .6616   & .7197 & .8133 \\ \midrule
    \multirow{4}{*}{Llama3-8B}
    & zero-shot                  & .3427   & .7099 & .8756 \\ 
    & 4-random                   & .5205   & .7339 & .8867 \\ 
    & 4-relation                 & .6871   & .7733 & .8934 \\ 
    & 4-template                 & .7268   & .7731 & .8628 \\ \midrule
    \multirow{4}{*}{Phi3-mini}
    & zero-shot                  & .3498   & .6346 & .8381 \\
    & 4-random                   & .4193   & .6506 & .8423 \\
    & 4-relation                 & .5686   & .6791 & .8436 \\
    & 4-template                 & .6067   & .6754 & .8114 \\ \midrule
    \multirow{4}{*}{Phi3-small}
    & zero-shot                  & .4258   & .6828 & .8642 \\
    & 4-random                   & .4889   & .7037 & .8695 \\
    & 4-relation                 & .6339   & .7172 & .8507 \\
    & 4-template                 & .6612   & .7181 & .8346 \\
    \bottomrule
    \end{tabular}
    \caption{Knowledge coverage rate on models with full-scale experiments. \textbf{Average}: mean accuracy using all templates; \textbf{Maximum}: accuracy with the best template for each relation; \textbf{Upper Limit}: accuracy when the best template is selected for each knowledge relation.}
    \label{table:full-scale-knowledge-coverage}
\end{table*}

\begin{table*}[t]
    \centering
    \small
    \begin{tabular}{lll}
    \toprule
    \textbf{ID} & 	\textbf{Human-rewritten templates} & 	\textbf{GPT-4 paraphrased templates} \\
    \midrule
    	\multirow{6}{*}[-3pt]{P19}
	&  \multirow{3}{*}{[X] started their life in [Y].}  & [X] took their first steps of life in [Y].  \\ 
	& & [X] activated their life's beginning in [Y]. \\
	& & [X] initiated their journey of life within [Y]. \\ \cmidrule{2-3}
	&  \multirow{3}{*}{The birth of [X] occurred in [Y].}  & The origin of [X] took place in [Y].  \\ 
	& & The inception of [X] was within [Y]. \\
	& & It was in [Y] that [X] first made its appearance. \\ \midrule
	\multirow{6}{*}[-3pt]{P20}
	&  \multirow{3}{*}{[X] spent the last years of life in [Y].}  & In [Y], [X] spent the end of their life.  \\ 
	& & [X]'s final era was in [Y]. \\
	& & In [Y], [X]'s life came to a close. \\ \cmidrule{2-3}
	&  \multirow{3}{*}{\makecell[l]{[Y] is the last place where [X]\\ lived until death.}}  & [X] inhabited [Y] up until death.  \\ 
	& & [Y] was the end-of-life dwelling for [X]. \\
	& & {[Y]} served as the last dwelling for {[X]} before they died. \\ \midrule
	\multirow{6}{*}[-3pt]{P279}
	&  \multirow{3}{*}{Of which class is [X] a subclass? [Y].}  & What is the general class that {[X]} is a part of as a subclass? {[Y]}.  \\ 
	& & What larger class encompasses {[X]} as a subclass? {[Y]}. \\
	& & Into which class is {[X]} categorized as a subclass? {[Y]}. \\ \cmidrule{2-3}
	&  \multirow{3}{*}{[X] is also necessarily a [Y].}  & [X] is intrinsically a [Y].  \\ 
	& & [X] is fundamentally a [Y]. \\
	& & [X] is by definition a [Y]. \\ \midrule
	\multirow{6}{*}[-3pt]{P37}
	&  \multirow{3}{*}{\makecell[l]{{[Y]} is spoken as an official language\\by people in {[X]}.}}  & {[Y]} is the authorized language for formal use in {[X]}. \\ 
	& & The official language spoken by individualsin {[X]} is {[Y]}. \\
	& & {[X]} endorses {[Y]} as the language forstate-related communication. \\ \cmidrule{2-3}
	&  \multirow{3}{*}{\makecell[l]{Officially, the people living in {[X]}\\use the language {[Y]} for communication.}}  & In {[X]}, the standard language for dialogue among the populace is {[Y]}.  \\ 
	& & Residents of [X] typically converse in [Y]. \\
	& & The official medium of verbal exchange in {[X]} is the {[Y]} language. \\ \midrule
	\multirow{6}{*}[-3pt]{P413}
	&  \multirow{3}{*}{[X] was given the [Y] job.}  & [X] was selected for the [Y] position.  \\ 
	& & [X] was named the new [Y]. \\
	& & The [Y] duties have been allocated to [X]. \\ \cmidrule{2-3}
	&  \multirow{3}{*}{[X] is a famous [Y] player.}  & {[X]} has risen to fame with their {[Y]} playing abilities.  \\ 
	& & [X] is well-known for playing [Y]. \\
	& & [X] is notable for their expertise in [Y]. \\ \midrule
	\multirow{6}{*}[-3pt]{P449}
	&  \multirow{3}{*}{[X] premiered on the network [Y].}  & [Y] was the origin of the broadcast for [X].  \\ 
	& & [X] was initially broadcasted by [Y]. \\
	& & The debut broadcast of [X] was on [Y]. \\ \cmidrule{2-3}
	&  \multirow{3}{*}{[Y] is the first air channel of [X].}  & [X] was originally brought to the public by [Y].  \\ 
	& & [X] first hit the airwaves courtesy of [Y]. \\
	& & [X] first reached listeners and viewers via [Y]. \\ \midrule
	\multirow{6}{*}[-3pt]{P47}
	&  \multirow{3}{*}{[X] and [Y] are neighboring countries.}  & {[X]} and {[Y]} are countries that are in close proximity.  \\ 
	& & [Y] lies in the vicinity of [X]. \\
	& & [Y] and [X] are countries that share a boundary. \\ \cmidrule{2-3}
	&  \multirow{3}{*}{You can go through [X] to reach [Y].}  & [X] acts as a gateway to [Y].  \\ 
	& & To reach [Y], one can travel through [X]. \\
	& & Traveling over [X] can bring you to [Y]. \\
    \bottomrule
    \end{tabular}
    \caption{Examples of templates in MyriadLAMA - part 1}
    \label{tab:template_example_1}
\end{table*}

\begin{table*}[t]
    \centering
    \small
\tabcolsep 1pt
    \begin{tabular}{lll}
    \toprule
    \textbf{ID} & 	\textbf{Human-rewritten templates} & 	\textbf{GPT-4 paraphrased templates} \\
    \midrule
    	\multirow{6}{*}[-3pt]{P138}
	&  \multirow{3}{*}{\makecell[l]{Who or what is [X] named \\ after? {[Y]}.}}  & Who is the namesake behind [X]? [Y].  \\ 
	& & What is the etymology behind [X]'s name? [Y]. \\
	& & Who or what was [X] called after? [Y]. \\ \cmidrule{2-3}
	&  \multirow{3}{*}{[X] is named after [Y].}  & [X] takes its name from [Y].  \\ 
	& & [Y] is the inspiration behind the name of [X]. \\
	& & [X] holds the name given in tribute to [Y]. \\ \midrule
	\multirow{6}{*}[-3pt]{P364}
	&  \multirow{3}{*}{[X] is created in language [Y].}  & [X] was composed in the [Y] language.  \\ 
	& & [X] unfolds in the language known as [Y]. \\
	& & [X] is expressed through the language of [Y]. \\ \cmidrule{2-3}
	&  \multirow{3}{*}{\makecell[l]{[X] was written in the [Y]\\ language.}}  & The words of [X] are in the [Y] language.  \\ 
	& & The composition of [X] is in the [Y] language. \\
	& & [X] was created using the [Y] language. \\ \midrule
	\multirow{6}{*}[-3pt]{P463}
	&  \multirow{3}{*}{[X] served for [Y].}  & [X] took part in [Y].  \\ 
	& & [X] collaborated with [Y]. \\
	& & [X] held a position at [Y]. \\ \cmidrule{2-3}
	&  \multirow{3}{*}{\makecell[l]{Which group or organization\\ does {[X]} belong to? {[Y]}.}}  & [X] is part of what organization? [Y].  \\ 
	& & [X] is a member of which entity? [Y]. \\
	& & Can you tell me which entity {[X]} is amember of? {[Y]}. \\ \midrule
	\multirow{6}{*}[-3pt]{P101}
	&  \multirow{3}{*}{\makecell[l]{Which field does [X] work in?\\ {[Y]}.}}  & In what industry is [X] employed? [Y].  \\ 
	& & [X] holds a position in which field? [Y]. \\
	& & [X] is a professional in what sector? [Y]. \\ \cmidrule{2-3}
	&  \multirow{3}{*}{\makecell[l]{[X] is influential in the domain\\ of [Y].}}  & The domain of {[Y]} feels the considerable impact of {[X]}.  \\ 
	& & [X] plays a pivotal role in the sphere of [Y]. \\
	& & [X] has a profound effect on [Y]. \\ \midrule
	\multirow{6}{*}[-3pt]{P106}
	&  \multirow{3}{*}{\makecell[l]{[X] is famous for\\ achievements as a [Y].}}  & {[X]} is well-known for their accomplishments in the {[Y]} role.  \\ 
	& & {[X]} is well-known for their successful career as a {[Y]}. \\
	& & {[X]} is a celebrated {[Y]} with a long list of achievements. \\ \cmidrule{2-3}
	&  \multirow{3}{*}{[X] is a [Y] by profession.}  & [X] has built a career as a [Y].  \\ 
	& & [X] is employed as a [Y]. \\
	& & [X] carries out the role of a [Y]. \\ \midrule
	\multirow{6}{*}[-3pt]{P527}
	&  \multirow{3}{*}{[Y] is a member of [X].}  & [X] contains [Y] as part of its composition.  \\ 
	& & [Y] holds a place in [X]. \\
	& & [Y] is a piece of [X]. \\ \cmidrule{2-3}
	&  \multirow{3}{*}{[Y] belongs to [X].}  & [Y] is held by [X].  \\ 
	& & [X] has [Y] under its ownership. \\
	& & [Y] is one of the items owned by [X]. \\ \midrule
	\multirow{6}{*}[-3pt]{P530}
	&  \multirow{3}{*}{\makecell[l]{{[Y]} is one of the countries {[X]}\\ has diplomatic relations with.}}  & {[Y]} is a member of the group of countries with which {[X]} conducts diplomacy.  \\ 
	& & {[X]} has a formal diplomatic relationship with {[Y]}, as it does with several other countries. \\
	& & {[Y]} is recognized by {[X]} as a diplomatic partner among other nations. \\ \cmidrule{2-3}
	&  \multirow{3}{*}{\makecell[l]{[X] has established diplomatic\\ ties with [Y].}}  & {[X]} has initiated formal diplomatic relations with {[Y]}.  \\ 
	& & [X] and [Y] have begun a diplomatic relationship. \\
	& & [X] and [Y] have set up official diplomatic links. \\

    \bottomrule
    \end{tabular}
    \caption{Examples of templates in MyriadLAMA - part 2}
    \label{tab:template_example_2}
\end{table*}

\begin{table*}[t]
    \centering
    \small
    \begin{tabular}{lll}
    \toprule
    \textbf{ID} & 	\textbf{Human-rewritten templates} & 	\textbf{GPT-4 paraphrased templates} \\
    \midrule
    	\multirow{6}{*}[-3pt]{P176}
	&  \multirow{3}{*}{[X] is a product of [Y]'s manufacturing.}  & The entity [Y] crafts and produces [X].  \\ 
	& & The item [X] is fabricated by [Y]. \\
	& & {[X]} is brought to life by {[Y]}'s manufacturing capabilities. \\ \cmidrule{2-3}
	&  \multirow{3}{*}{Which company produced [X]? [Y].}  & Can you tell me who made [X]? [Y].  \\ 
	& & Which producer can be linked to [X]? [Y]. \\
	& & What is the producing company of [X]? [Y]. \\ \midrule
	\multirow{6}{*}[-3pt]{P27}
	&  \multirow{3}{*}{[X] is a person from country [Y].}  & [X] is a resident of [Y].  \\ 
	& & [X] bears the nationality of [Y]. \\
	& & [X] is a product of [Y]. \\ \cmidrule{2-3}
	&  \multirow{3}{*}{The nationality of [X] is [Y].}  & [X] is a native of [Y].  \\ 
	& & [X] is identified as a national from [Y]. \\
	& & [Y] is the country of origin for [X]. \\ \midrule
	\multirow{6}{*}[-3pt]{P407}
	&  \multirow{3}{*}{[X] is in language [Y].}  & The language of [X] is [Y].  \\ 
	& & The primary linguistic expression of {[X]} is in {[Y]}. \\
	& & [X] is articulated through the [Y] language. \\ \cmidrule{2-3}
	&  \multirow{3}{*}{[X] is a work in the [Y] language.}  & The [Y] language is the linguistic fabric of [X].  \\ 
	& & [X] has been produced using the [Y] language. \\
	& & {[X]} is an example of literature in the {[Y]} language. \\ \midrule
	\multirow{6}{*}[-3pt]{P30}
	&  \multirow{3}{*}{On what continent is [X] located? [Y].}  & What's the name of the continent that {[X]} calls home? {[Y]}.  \\ 
	& & What continental landmass does [X] occupy? [Y]. \\
	& & [X] lies on which of the Earth's continents? [Y]. \\ \cmidrule{2-3}
	&  \multirow{3}{*}{[X] is a part of the continent [Y].}  & [X] is a section of the continental land of [Y].  \\ 
	& & {[X]} is geographically positioned as part of continent {[Y]}. \\
	& & [X] is an integral piece of the continent [Y]. \\ \midrule
	\multirow{6}{*}[-3pt]{P178}
	&  \multirow{3}{*}{[X] was originally created by [Y].}  & The foundation of [X] was laid by [Y].  \\ 
	& & The concept of [X] was conceived by [Y]. \\
	& & [X] first came into existence thanks to [Y]. \\ \cmidrule{2-3}
	&  \multirow{3}{*}{[X] is developed by [Y].}  & [Y] has developed [X].  \\ 
	& & [Y] is the developer behind [X]. \\
	& & [Y] stands as the creator of [X]. \\ \midrule
	\multirow{6}{*}[-3pt]{P1376}
	&  \multirow{3}{*}{[X] is the capital of [Y].}  & [Y]'s governmental seat is in [X].  \\ 
	& & [X] is recognized as the official capital of [Y]. \\
	& & The leading city and capital of [Y] is [X]. \\ \cmidrule{2-3}
	&  \multirow{3}{*}{[X] is the administrative center of [Y].}  & {[Y]}'s administrative leadership is situated in {[X]}.  \\ 
	& & [Y]'s administrative affairs are managed from [X]. \\
	& & {[X]} is where {[Y]}'s administrative management is anchored. \\ \midrule
	\multirow{6}{*}[-3pt]{P131}
	&  \multirow{3}{*}{[Y] is the place where [X] is located.}  & [X] resides in [Y].  \\ 
	& & [X] can be found at the location of [Y]. \\
	& & [X] is anchored in [Y]. \\ \cmidrule{2-3}
	&  \multirow{3}{*}{[X] is located in [Y].}  & [Y] is where [X] is established.  \\ 
	& & [Y] contains [X]. \\
	& & [Y] houses [X]. \\

    \bottomrule
    \end{tabular}
    \caption{Examples of templates in MyriadLAMA - part 3}
    \label{tab:template_example_3}
\end{table*}

\begin{table*}[t]
    \centering
    \small
    \begin{tabular}{lll}
    \toprule
    \textbf{ID} & 	\textbf{Human-rewritten templates} & 	\textbf{GPT-4 paraphrased templates} \\
    \midrule
    	\multirow{6}{*}[-3pt]{P1412}
	&  \multirow{3}{*}{What language does [X] use? [Y].}  & [X] communicates in what vernacular? [Y].  \\ 
	& & What tongue does [X] utilize? [Y]. \\
	& & What is the primary language for [X]? [Y]. \\ \cmidrule{2-3}
	&  \multirow{3}{*}{[Y] is the language that is used by [X].}  & The tongue of [X] is the language [Y].  \\ 
	& & [X] uses [Y] as its mode of speech. \\
	& & {[Y]} is the language that enables communication for {[X]}. \\ \midrule
	\multirow{6}{*}[-3pt]{P108}
	&  \multirow{3}{*}{[X] is employed by [Y].}  & [Y] is the employer of [X].  \\ 
	& & [X] has a job at [Y]. \\
	& & [Y] is the source of employment for [X]. \\ \cmidrule{2-3}
	&  \multirow{3}{*}{Who does [X] work for? [Y].}  & Who does [X] report to in their job? [Y].  \\ 
	& & For whom is [X] currently working? [Y]. \\
	& & Who holds [X] on their team? [Y]. \\ \midrule
	\multirow{6}{*}[-3pt]{P136}
	&  \multirow{3}{*}{What is the genre of [X]? [Y].}  & \makecell[l]{In terms of genre, how would you classify\\{[X]}? {[Y]}.}  \\ 
	& & What category of genre does [X] belong to? [Y]. \\
	& & In what genre category would you place [X]? [Y]. \\ \cmidrule{2-3}
	&  \multirow{3}{*}{[X] is the representative of the [Y] style.}  & [X] personifies the [Y] style in its purest form.  \\ 
	& & [X] is the epitome of the [Y] approach. \\
	& & [X] is the archetype of the [Y] tradition. \\ \midrule
	\multirow{6}{*}[-3pt]{P17}
	&  \multirow{3}{*}{Which country is [X] located? [Y].}  & Can you identify the country where {[X]} is situated? {[Y]}.  \\ 
	& & Could you specify the country of {[X]}'s location? {[Y]}. \\
	& & [X] can be located in what country? [Y]. \\ \cmidrule{2-3}
	&  \multirow{3}{*}{[Y] is the country in which [X] is located.}  & [Y] is the nation that houses [X].  \\ 
	& & [Y] encompasses the region where [X] can be found. \\
	& & [Y] is the setting for the location of [X]. \\ \midrule
	\multirow{6}{*}[-3pt]{P39}
	&  \multirow{3}{*}{What position does [X] hold? [Y].}  & What position does [X] occupy? [Y].  \\ 
	& & What is the employment status of [X]? [Y]. \\
	& & What is the position title for [X]? [Y]. \\ \cmidrule{2-3}
	&  \multirow{3}{*}{[X] was sworn in as [Y].}  & [X] has been designated the official role of [Y].  \\ 
	& & [X] pledged their commitment to the role of [Y]. \\
	& & [X] was confirmed in the role of [Y]. \\ \midrule
	\multirow{6}{*}[-3pt]{P264}
	&  \multirow{3}{*}{Which music label represents [X]? [Y].}  & Which label has [X] on its roster? [Y].  \\ 
	& & Who is [X]'s music label? [Y]. \\
	& & With whom is [X] signed for music production? [Y]. \\ \cmidrule{2-3}
	&  \multirow{3}{*}{[X] is represented by music label [Y].}  & The music label acting on behalf of [X] is [Y].  \\ 
	& & [Y] is the music label that has signed [X]. \\
	& & [X] has music label [Y] as its representative. \\ \midrule
	\multirow{6}{*}[-3pt]{P276}
	&  \multirow{3}{*}{Where is [X] located? [Y].}  & What's the location of [X]? [Y].  \\ 
	& & Where can [X] be found? [Y]. \\
	& & Where should I look for [X]? [Y]. \\ \cmidrule{2-3}
	&  \multirow{3}{*}{[X] is located in [Y].}  & [X] is positioned in [Y].  \\ 
	& & [X] occupies a space in [Y]. \\
	& & [Y] contains [X]. \\

    \bottomrule
    \end{tabular}
    \caption{Examples of templates in MyriadLAMA - part 4}
    \label{tab:template_example_4}
\end{table*}

\begin{table*}[t]
    \centering
    \small
\tabcolsep2.9pt
    \begin{tabular}{lll}
    \toprule
    \textbf{ID} & 	\textbf{Human-rewritten templates} & 	\textbf{GPT-4 paraphrased templates} \\
    \midrule
    	\multirow{6}{*}[-3pt]{P937}
	&  \multirow{3}{*}{[Y] is the place where [X] worked.}  & [X] had their employment based in [Y].  \\ 
	& & [X] found their employment setting in [Y]. \\
	& & {[X]} conducted their professional activities in {[Y]}. \\ \cmidrule{2-3}
	&  \multirow{3}{*}{[X] had work activity in [Y].}  & [X] took part in business tasks in [Y].  \\ 
	& & [X] was employed within the confines of [Y]. \\
	& & [X] was operational in the workforce at [Y]. \\ \midrule
	\multirow{6}{*}[-3pt]{P140}
	&  \multirow{3}{*}{Which religion is [X] affiliated with? [Y].}  & What religious belief does [X] adhere to? [Y].  \\ 
	& & Which spiritual path is embraced by [X]? [Y]. \\
	& & What is the creed of [X]? [Y]. \\ \cmidrule{2-3}
	&  \multirow{3}{*}{[X] is affiliated with the [Y] religion.}  & [X] is part of the [Y] religious denomination.  \\ 
	& & {[X]} is associated with the {[Y]} spiritual tradition. \\
	& & [X] adheres to the [Y] religion. \\ \midrule
	\multirow{6}{*}[-3pt]{P1303}
	&  \multirow{3}{*}{[X] is a [Y] player.}  & [X] specializes in the [Y].  \\ 
	& & [X] is a seasoned [Y] player. \\
	& & [X] is a [Y] specialist. \\ \cmidrule{2-3}
	&  \multirow{3}{*}{[X] plays [Y].}  & [X] expresses their musicianship through [Y].  \\ 
	& & [X] has chosen [Y] as their musical companion. \\
	& & [X] is a musician who specializes in [Y]. \\ \midrule
	\multirow{6}{*}[-3pt]{P127}
	&  \multirow{3}{*}{Who owns [X]? [Y].}  & Whose property is [X] considered to be? [Y].  \\ 
	& & Who is the legal holder of [X]? [Y]. \\
	& & Who has the ownership rights to [X]? [Y]. \\ \cmidrule{2-3}
	&  \multirow{3}{*}{[X] is owned by [Y].}  & [Y] is the proprietor of [X].  \\ 
	& & [Y] holds the title to [X]. \\
	& & [Y] possesses [X]. \\ \midrule
	\multirow{6}{*}[-3pt]{P103}
	&  \multirow{3}{*}{\makecell[l]{[X] grew up speaking [Y] as their first\\ language.}}  & [X]’s formative years were shaped by speaking [Y].  \\ 
	& & [X] started their life speaking [Y]. \\
	& & [X]’s childhood language was [Y]. \\ \cmidrule{2-3}
	&  \multirow{3}{*}{[Y] is the mother tongue of [X].}  & [X] has [Y] as their original tongue.  \\ 
	& & {[X]} was nurtured in an environment where {[Y]} is spoken. \\
	& & [X] has [Y] as the language of their upbringing. \\ \midrule
	\multirow{6}{*}[-3pt]{P190}
	&  \multirow{3}{*}{The city of [X] is twinned with [Y].}  & {[Y]} and {[X]} have entered into a twinning arrangement.  \\ 
	& & [X] is in a twinning relationship with [Y]. \\
	& & A twinning link has been established between {[X]} and {[Y]}. \\ \cmidrule{2-3}
	&  \multirow{3}{*}{\makecell[l]{{[X]} and {[Y]} are sister cities that have\\ been developing together.}}  & {[X]} and {[Y]} have been sister cities on a shared developmental journey.  \\ 
	& & The cities of {[X]} and {[Y]} have jointly progressed as sister municipalities. \\
	& & {[X]} and {[Y]} have been in lockstep as sister cities in their development. \\ \midrule
	\multirow{6}{*}[-3pt]{P1001}
	&  \multirow{3}{*}{[X] applies to the jurisdiction in [Y].}  & The jurisdiction of [Y] encompasses [X].  \\ 
	& & [X] is answerable to the legal system in [Y]. \\
	& & [Y] exercises legal control over [X]. \\ \cmidrule{2-3}
	&  \multirow{3}{*}{The region of [Y] uses [X] as a legal term.}  & [X] is a term with legal standing in [Y].  \\ 
	& & The legal system of {[Y]} includes {[X]} as an official term. \\
	& & [X] is employed as a juridical term in [Y]. \\

    \bottomrule
    \end{tabular}
    \caption{Examples of templates in MyriadLAMA - part 5}
    \label{tab:template_example_5}
\end{table*}

\begin{table*}[t]
    \centering
    \small
    \begin{tabular}{lll}
    \toprule
    \textbf{ID} & 	\textbf{Human-rewritten templates} & 	\textbf{GPT-4 paraphrased templates} \\
    \midrule
    	\multirow{6}{*}[-3pt]{P31}
	&  \multirow{3}{*}{[X] is a [Y].}  & [X] represents a [Y].  \\ 
	& & [X] is an example of a [Y]. \\
	& & [X] is termed a [Y]. \\ \cmidrule{2-3}
	&  \multirow{3}{*}{Speaking of [Y], [X] is an example of it.}  & [X] is a particular instance that reflects [Y].  \\ 
	& & {[X]} is a variant that falls within the scope of {[Y]}. \\
	& & [Y] can be demonstrated through [X]. \\ \midrule
	\multirow{6}{*}[-3pt]{P495}
	&  \multirow{3}{*}{[X] originates from the country of [Y].}  & [X] was first found in the land of [Y].  \\ 
	& & The inception of [X] is linked to the country [Y]. \\
	& & The origin of [X] can be traced back to [Y]. \\ \cmidrule{2-3}
	&  \multirow{3}{*}{[X] first appeared in [Y].}  & [X] has its roots in [Y].  \\ 
	& & [X] was first crafted in [Y]. \\
	& & The origin of [X] is attributed to [Y]. \\ \midrule
	\multirow{6}{*}[-3pt]{P159}
	&  \multirow{3}{*}{\makecell[l]{The operation of {[X]} depends on the\\ headquarters in {[Y]}.}}  & {[X]}'s functioning is reliant on the main office in {[Y]}.  \\ 
	& & The base in [Y] is essential for [X] to function. \\
	& & The primary operations of {[X]} are contingent upon the base in {[Y]}. \\ \cmidrule{2-3}
	&  \multirow{3}{*}{The headquarters of [X] is in [Y].}  & [Y] is home to the central office of [X].  \\ 
	& & The top office of [X] is positioned in [Y]. \\
	& & The nerve center for {[X]}'s operations is based in {[Y]}. \\ \midrule
	\multirow{6}{*}[-3pt]{P36}
	&  \multirow{3}{*}{[Y] is the administrative center of [X].}  & The nerve center for {[X]}'s administration is found in {[Y]}.  \\ 
	& & {[X]}'s administrative governance is centralized in {[Y]}. \\
	& & [X] is administratively governed by [Y]. \\ \cmidrule{2-3}
	&  \multirow{3}{*}{[Y] represents the capital city for [X].}  & [Y] functions as [X]'s political hub.  \\ 
	& & [X] uses [Y] as its head city. \\
	& & [X]'s administrative center is [Y]. \\ \midrule
	\multirow{6}{*}[-3pt]{P740}
	&  \multirow{3}{*}{[X] started their career in [Y].}  & [Y] served as the starting point for [X]'s career.  \\ 
	& & {[X]} began earning their stripes in the field of {[Y]}. \\
	& & [X] commenced their employment journey with [Y]. \\ \cmidrule{2-3}
	&  \multirow{3}{*}{The formation location of [X] is [Y].}  & The assembly point for [X] is [Y].  \\ 
	& & {[Y]} is recognized as the setting for {[X]}'s formation. \\
	& & [Y] is where [X] originates. \\ \midrule
	\multirow{6}{*}[-3pt]{P361}
	&  \multirow{3}{*}{Which entity does [X] belong to? [Y].}  & Who owns [X]? [Y].  \\ 
	& & What is the overarching group for [X]? [Y]. \\
	& & What organization encompasses [X]? [Y]. \\ \cmidrule{2-3}
	&  \multirow{3}{*}{[Y] consists of [X].}  & [X] is what [Y] is primarily made of.  \\ 
	& & [Y] incorporates [X] within it. \\
	& & [Y] is structured with [X]. \\
    \bottomrule
    \end{tabular}
    \caption{Examples of templates in MyriadLAMA - part 6}
    \label{tab:template_example_6}
\end{table*}

\end{document}